\documentclass[letterpaper]{article} 
\usepackage[draft]{aaai2027}  
\usepackage{times}  
\usepackage{helvet}  
\usepackage{courier}  
\usepackage[hyphens]{url}  
\usepackage{graphicx} 
\usepackage{natbib}  
\usepackage{caption} 
\usepackage{amsmath,amssymb,amsfonts}
\usepackage{booktabs}
\usepackage{array}
\usepackage{tabularx}

\newcommand{\R}{\mathbb{R}}

\newcommand{\KL}{\mathrm{KL}}
\newcommand{\DJS}{D_{\mathrm{JS}}}
\newcommand{\Cat}{\mathrm{Cat}}
\newcommand{\W}{\mathcal{W}}
\newcolumntype{Y}{>{\raggedright\arraybackslash}X}

\title{Graph Feedback Controls Consensus and Clique Formation in Open-Weight Language-Model Populations}
\author{
    Samer Saab Jr. \qquad  Chaouki Abdallah
}
\affiliations{
    Electrical and Computer Engineering Department\\
    Lebanese American University\\
    Byblos, Lebanon
}

\begin{document}
\maketitle

\begin{abstract}
Multi-agent language-model (LM) systems often determine which agents communicate, yet routing is usually treated as an implementation detail. We ask whether routing itself determines whether a population converges on a shared convention or fragments into persistent cliques. We study open-weight agents spanning 1.1B–32B parameters in a controlled naming game, tracking both emitted labels and full first-token preference distributions over the allowed labels. Similarity-based routing can isolate emerging conventions and sustain fragmentation even when every agent interacts in every round. Matched controls show that this effect is not explained solely by uneven participation or model-family-specific score preferences: random rematching and policies that connect disagreeing groups improve coordination when partner-label history is retained, but not when it is absent. Exposure alone is nevertheless insufficient, as some mixed-model populations remain divided despite frequent cross-family interaction, although the same models coordinate homogeneously. Trajectory and controlled-history analyses further distinguish reaching consensus from maintaining it. Finally, ARC-Challenge and MMLU experiments show that routing changes how correct and incorrect answers propagate without reliably improving accuracy. These results establish the runtime interaction graph as a causal design variable whose effects depend jointly on memory, model response, and population composition.
\end{abstract}

\section{Introduction}
Multi-agent language-model (LM) systems increasingly rely on local communication; agents are paired, clustered, routed, or allowed to exchange partial histories before producing collective outputs. This paper tests a simple design claim: the runtime interaction graph is causal. A routing rule can keep emerging groups in contact and allow disagreement to dissipate, or it can isolate those groups and stabilize persistent fragmentation.

We use a controlled naming game. Each of $N$ agents repeatedly chooses one artificial label from a fixed set. The labels are selected to be single tokens for every model in the population, so agreement reflects social coordination rather than recovery of a semantically correct answer. The agents change only through their prompts, retained interaction histories, and sampling; model parameters remain fixed.

Open-weight models expose a score for every allowed next-token label. Normalizing these scores gives each agent a measured preference distribution over the label set, which we call its score state. This lets us distinguish behavioral consensus, where all agents emit the same label, from score-state connectedness, where the agents' preference distributions form one connected similarity graph. A connected graph does not imply that the distributions are nearly identical, so we also report pairwise score distances and total score-state dispersion. We use convention basin to mean a group of agents with similar score-state preferences, and bridge to mean an interaction connecting agents from different basins.

The experiments vary four elements of the collective system: who interacts, what social history agents retain, which model families populate the group, and which score representation is used to select partners. We study TinyLlama-1.1B-Chat, Yi-1.5-6B-Chat, Qwen2-7B-Instruct, and Qwen2.5-Instruct models at 7B, 14B, and 32B scales.
Across homogeneous and heterogeneous populations, the experiments isolate the effects of memory, routing, model family, scale, population composition, and stability after consensus. Matched mixed-population substitutions test whether evidence exchanged across model families is actually adopted, while prior-normalized routing separates each model family's default score pattern from changes produced during interaction. Controlled prompts measure how individual models respond to the same partner-label records, and ARC-Challenge and MMLU experiments test whether the same propagation mechanisms extend from artificial conventions to real answer choices.

\begin{figure*}[t]
    \centering
    \includegraphics[width=0.85\textwidth]{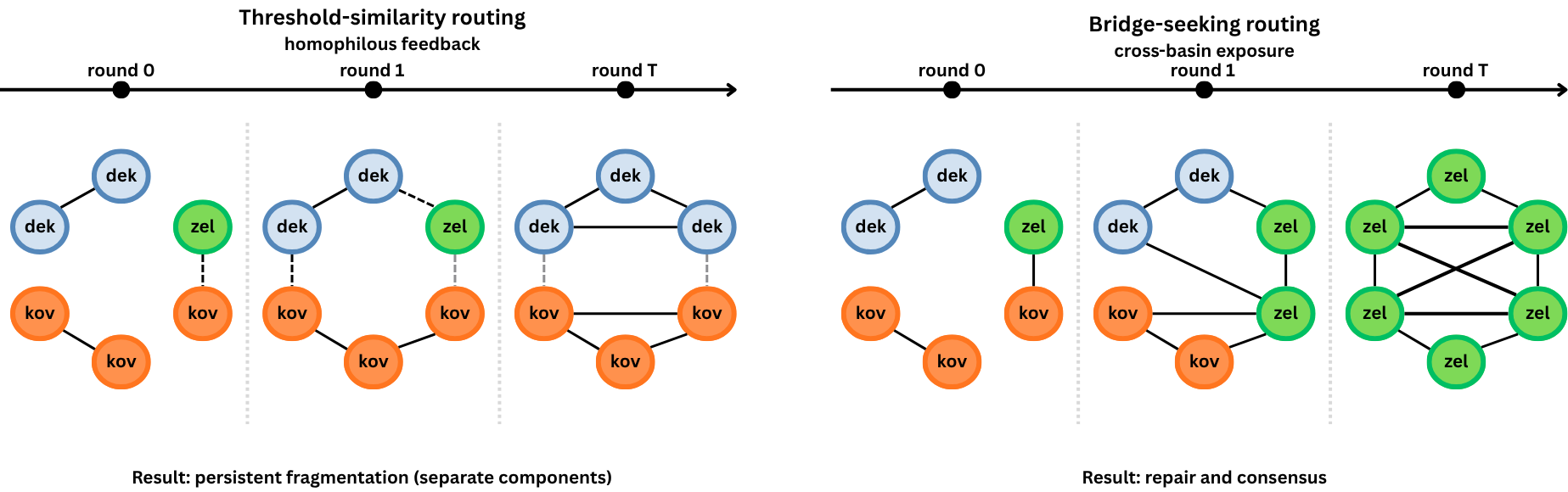}
    \caption{Schematic comparison of similarity-threshold and bridge routing. Left: future partners are eligible only when their previous-round score states are sufficiently similar. As two groups separate, interactions between them disappear and each group continues internally. Right: bridge policies deliberately connect agents from different score-state groups or with different sampled labels, allowing partner-label evidence to move between groups. Whether this exposure produces lasting agreement also depends on how the models respond to the evidence they receive.}
    \label{fig:schematic}
\end{figure*}

Our contributions are threefold. First, we measure both what agents emit and how their full label-preference distributions relate, using label graphs, score-state graphs constructed from Jensen--Shannon distance, and measures of population disagreement. Second, controls over partner choice and participation show that randomizing partners improves coordination under the same participation schedule, redistributing the same interaction budget improves it further, and similarity-based matching can preserve fragmentation even when every agent interacts in every round. Prior-normalized routing tests whether mixed-population fragmentation is explained only by each model family's default score pattern. The matched Qwen2.5-14B/32B comparison separately distinguishes reaching
consensus from maintaining it. Third, matched changes in population composition and controlled response probes show why access to evidence is not sufficient: an agent may observe another model family's convention without adopting and retaining it. ARC-Challenge and MMLU experiments extend this propagation analysis to real answer choices.
Figure~\ref{fig:schematic} summarizes the central mechanism.

\section{Related Work}
Convention emergence and social norms have long been studied as decentralized coordination problems in which repeated local interaction selects among multiple equivalent equilibria \cite{young1996economics,shoham1997emergence,delgado2002emergence,sen2007emergence,airiau2014emergence}. Naming games provide a minimal protocol for shared vocabulary formation, metastability, and tipping phenomena \cite{baronchelli2006sharp,dallasta2006nonequilibrium,centola2018experimental}. We use nonce labels for the same reason: there is no externally correct convention, so agreement reflects social propagation rather than semantic recovery.

Consensus and opinion-dynamics models supply the graph language used in this paper. DeGroot-style averaging, nearest-neighbor consensus, and switching-topology consensus identify connectivity and repeated exposure as structural conditions for global agreement \cite{degroot1974reaching,jadbabaie2003coordination,olfati2004consensus,moreau2005stability,ren2005consensus}. Adaptive-network and bounded-confidence models show the complementary failure mode: when interaction depends on current state, the system can polarize or settle into cluster consensus rather than global consensus \cite{hegselmann2002opinion,holme2006nonequilibrium,gross2008adaptive,lipowska2012naming}. We use this theory diagnostically, not literally. LM agents update through prompts, retained histories, priors, and sampling, so graph connectivity is an opportunity for repair rather than a guarantee of linear averaging.

Recent work on LM-agent populations studies social conventions, collective bias, opinion dynamics, norm-like behavior, and multi-agent collaboration mechanisms \cite{ashery2025emergent,chuang2024simulating,ren2024emergence,horiguchi2024evolution,takata2024spontaneous,tran2025multiagent, saab2026prompts}. The closest connection is the observation that LM populations can form conventions or shared biases. Our question is more structural: which runtime exposure graphs make agreement, fragmentation, or residual disagreement in score space more likely? The open-weight setting lets us answer at both the behavioral level, through sampled labels, and the score-state level, through restricted first-token distributions over the allowed labels.

\section{Model and Diagnostics}
\paragraph{Naming game.}
At each round $t$, each agent outputs one label from a finite convention set $\W=\{w_1,\ldots,w_M\}$. The labels are artificial nonce strings, so agreement indicates convention formation rather than recovery of a semantically correct answer. At the beginning of a round, the runtime controller selects a matching over agents. A matching is a set of disjoint pairs, so each agent interacts with at most one partner in that round. Both agents in a matched pair are queried before either current-round output is added to any prompt, so current-round communication is not sequential.

Agent prompts contain the coordination instruction, the allowed-label set, the last and cumulative coordination rewards, and at most $H$ recent local interaction records. Each record shows the agent's previous label, its partner's label, whether they matched, and the resulting binary reward. The executed prompt does not display the round index. Thus, $H=0$ replaces the history with an empty-history placeholder, while $H=3$ and $H=10$ provide short retained memory. The allowed-label order is reshuffled for every query. We use ``no-memory'' to mean no rendered partner-label records, not absence of all prompt-level variation.

\paragraph{Measured state.}
For open-weight models, let $z_i(t)\in\R^M$ contain the model's unnormalized first-token scores, or logits, restricted to the $M$ allowed labels. We choose labels that are single tokens for every tokenizer in the cohort. The measured score-state distribution is the temperature-one restricted softmax
\begin{equation}
  s_i(t)_m=\frac{\exp(z_{im}(t))}{\sum_{\ell=1}^M \exp(z_{i\ell}(t))},\quad s_i(t)\in\Delta^{M-1}.
\end{equation}
The emitted label is sampled using the run's decoding temperature $\tau$,
\begin{equation}
  y_i(t)\sim\Cat\left(\frac{\exp(z_i(t)/\tau)}{\sum_{\ell=1}^M \exp(z_{i\ell}(t)/\tau)}\right).
\end{equation}
Unless stated otherwise, ``state'' refers to $s_i(t)$, not the sampling distribution. State distance is metric Jensen-Shannon distance,
\begin{align}
\DJS(p,q)&=\left[\tfrac12\KL(p\Vert m)+\tfrac12\KL(q\Vert m)\right]^{1/2},\notag\\
m&=\tfrac12(p+q).
\end{align}

\paragraph{Graph views and energies.}
We distinguish four graphs. The exposure graph records who interacted. The label graph connects agents with the same sampled label. The state-similarity graph connects agents when $\DJS(s_i(t),s_j(t))\leq\Delta$. The endogenous runtime graph is the graph used to choose future pairings. For a representation $u_i(t)$, define
\begin{equation}
  V_u(t)=\frac12\sum_{i=1}^N\lVert u_i(t)-\bar u(t)\rVert_2^2,
\end{equation}
where $u_i$ is either $s_i(t)$ or a one-hot sampled label. The algebraic connectivity $\lambda_2(L(t))$ of a diagnostic or runtime graph indicates whether global bridges exist. For an undirected graph, $\lambda_2>0$ exactly when the graph is connected; larger values indicate stronger global connectivity. These quantities are not assumed to be the LM update rule; rather, they are observables inspired by switched-consensus dynamics. In the reference model $\dot{x}=-(L(t)\otimes I_M)x+r(t)$, positive repeated connectivity creates an opportunity for disagreement dissipation, whereas persistent disconnection permits cluster consensus.

\paragraph{Why homophilous routing can disconnect basins.}
Intuitively, if two groups are farther apart in score space than the routing threshold, even after accounting for variation within each group, the threshold controller cannot create an edge between them. If agents remain within radii $B_m$ and $B_n$ of score-state
prototypes $\mu_m$ and $\mu_n$, then the triangle inequality gives
\[
D_{\mathrm{JS}}(s_i,s_j)
\geq
D_{\mathrm{JS}}(\mu_m,\mu_n)-B_m-B_n.
\]
Thus, when the distance between the two group prototypes exceeds $\Delta+B_m+B_n$, no threshold edge can connect the groups. Once the runtime graph separates into components, exposure no longer forces their conventions to interact; the full argument is given in Supplementary Sec.~I.4.

\paragraph{Runtime controllers.}
Well-mixed routing samples a random matching from the complete agent graph, so every pair is eligible before the matching is drawn. The runtime controller constructs a candidate graph before querying the models in round $t$. At $t=0$, no previous observations exist, so the controller initializes from the complete graph without self-loops. For $t\geq 1$, the candidate graph is computed only from observations saved at round $t-1$; the controller has no access to current-round labels or current-round score states.

The threshold controller is homophilous, as it builds the next candidate graph from previous state similarity,
\[
a^{\mathrm{end}}_{ij}(t)=\mathbf{1}\{\DJS(s_i(t-1),s_j(t-1))\leq\Delta\}.
\]
To obtain disjoint pairs, the controller sorts eligible edges by weight, breaks near-ties randomly, and adds an edge whenever neither endpoint has already been paired. It stops when no further eligible pair can be added. This is a greedy maximal matching. Threshold-similarity uses this maximal matching without a pairing fallback, so a sparse thresholded candidate graph may leave agents unmatched. Bridge controllers first select pairs that connect disagreeing groups. Any agents left unmatched are then paired randomly when possible; we call this well-mixed fallback. State-component bridge routing pairs agents across previous state-similarity components. Label-disagreement routing pairs agents whose previous sampled labels differ. State-distance routing prioritizes pairs whose previous score states are separated, but not maximally different, using the distance band $0.05\leq\DJS\leq0.50$. Component-bridge runs should therefore be interpreted as bridge routing with fallback, not as a pure cross-component-only process.

\paragraph{Prior-normalized routing state.}
Different model families can have different default label preferences even before they observe any social history. We estimate this empty-history preference distribution separately for each family. Before computing routing distances, we remove that family-specific baseline from the agent's current score state and renormalize the result. Agents still sample labels from their unadjusted logits, so normalization changes who can be paired rather than directly changing what a model outputs. The estimator and the raw-versus-normalized diagnostics are detailed in Supplementary Sec.~B.1.4.

\paragraph{Controls for partner choice and participation.}
Threshold routing determines both which agents participate and which participating agents meet. We use three controls to separate these effects. For each threshold trajectory, let $A_t$ be the agents who participate in round $t$ and let $k_t=|A_t|/2$ be the number of realized pairs. The active-set-matched control randomly rematches the agents in $A_t$, preserving the exact participants, pair count, per-agent participation totals, and unpaired schedule while changing their partners. The pair-count-matched control preserves only $k_t$: it randomly selects $2k_t$ agents and randomly matches them, redistributing participation while holding the interaction budget fixed.

The full-coverage similarity control instead pairs all ten agents in every round and selects the perfect matching with the smallest total previous-round score-state distance. It therefore removes unequal participation while preserving a strong preference for similar partners. The two schedule-matched controls each contain 189 trajectories, while the full-coverage condition contains 27 trajectories over all history, temperature, and seed settings. All controls inherit the corresponding model ordering, prompts, labels, decoding settings, initialization, and reward bookkeeping, and every trajectory runs for 200 rounds.

\section{Experiments}
The experimental suite contains coordinated contrasts over routing, memory, model family, scale, and population composition. The heterogeneous controller grid uses $N=10$ agents drawn from TinyLlama-1.1B-Chat, Yi-1.5-6B-Chat, Qwen2-7B-Instruct, and Qwen2.5-7B-Instruct. The fixed backend cycle is Qwen2.5-7B, Qwen2-7B, Yi-6B, and TinyLlama, giving multiplicities $3,3,2,2$. Homogeneous populations span TinyLlama-1.1B, Yi-1.5-6B, and Qwen2.5 at 7B, 14B, and 32B under the same $N=10$, $M=10$, $T=200$, history, temperature, and state-threshold design. TinyLlama, Yi, Qwen2.5-14B, and Qwen2.5-32B are evaluated under complete well-mixed, threshold-similarity, state-component, label-disagreement, and state-distance controller grids; Qwen2.5-7B supplies the matched history sweep and targeted controller contrasts. The prior-normalized mixed suite uses the same 189-trajectory threshold design and the corresponding 27-trajectory state-component and state-distance designs, with calibration-adjusted states used for routing and raw restricted logits retained for output sampling. Unless stated otherwise, $\Delta=0.25$, $\tau\in\{0.1,0.3,0.7\}$, and $H\in\{0,3,10\}$.

The population-composition contrast contains five $N=10$ cohorts. Two populations share the same 32B/14B/7B Qwen backbone and differ only in whether two positions are occupied by TinyLlama or Yi. A fifth population removes the 14B agents while retaining Qwen2.5-32B, Qwen2.5-7B, Yi, and TinyLlama. All five cohorts use $H\in\{0,3\}$, $\tau=0.3$, $\Delta=0.25$, and seeds 11--13 under well-mixed, threshold-similarity, state-component, label-disagreement, and state-distance routing, giving 150 runs. The attainment-and-persistence comparison uses all 594 homogeneous 14B and 32B trajectories to form 297 matched setting pairs, with 72 retained-memory non-threshold settings per model in the principal comparison. Controlled response probes use three tokenizer-safe vocabularies, randomized label orders, consistent and mixed three-record histories, and seeded population controls. The task-grounded branch uses Qwen2.5-7B agents on 295 fixed-four-choice ARC-Challenge validation items and 100 MMLU questions.

Open-weight runs use tokenizer-safe nonce labels chosen to be single tokens for every tokenizer in the relevant cohort. After all agents have sampled labels in round $t$, partner-label feedback and local histories are updated, and the saved observations from round $t$ may be used to construct the candidate graph for round $t+1$. This ordering is important for causal interpretation. That is, runtime graph feedback is based on previous-round observations, not on current-round outcomes.

A round has full behavioral consensus when all agents emit the same allowed label. Final behavioral consensus applies this criterion in the last round. Stable final-20 behavioral consensus requires it in every one of the final 20 rounds. Transient behavioral consensus means that it occurs in at least one round, even if it is later lost.
Final score-state connectedness means that the final score-state similarity graph has one connected component. Final match is the fraction of unordered agent pairs whose final sampled labels agree. Score-state connectedness is a binary graph property, while pairwise Jensen--Shannon distances and disagreement energy measure how tightly the score distributions have contracted.
Each reported trajectory corresponds to one experimental configuration and one random seed. History, temperature, threshold, controller, or population changes define a new configuration; seeds provide repeated trajectories within that configuration.

\section{Results}
\paragraph{Memory separates fragmentation from convention formation.}

The history sweep isolates whether social evidence can accumulate under otherwise well-mixed exposure. None of the 45 no-memory runs across TinyLlama, Yi, and Qwen2.5 at 7B, 14B, and 32B reaches final behavioral consensus or final score-state connectedness. With retained history, every model reaches full behavioral consensus at least transiently in all 18 runs, but terminal outcomes are model conditioned. Qwen2.5-7B and Qwen2.5-32B reach final behavioral consensus and score-state connectedness in 18/18 runs. Yi reaches final behavioral consensus in 17/18 and score-state connectedness in 18/18; TinyLlama reaches the two endpoints in 17/18 and 11/18; and Qwen2.5-14B reaches both in 6/18. TinyLlama at $H=3$ already reaches final behavioral consensus in 8/9 runs but score-state connectedness in only 2/9, whereas $H=10$ yields 9/9 for both. Retained history therefore helps every model leave the fragmented regime, but the model determines whether final label agreement is accompanied by connected score states and whether that agreement persists.

\begin{table}[h]
\caption{Homogeneous well-mixed history sweep across five open-weight models. Each row pools seeds 11--13 and temperatures $\tau\in\{0.1,0.3,0.7\}$, giving nine runs per history horizon. ``Final label / state conn.'' reports final behavioral-consensus and final score-state-connectedness counts.}
\label{tab:history}
\centering
\tiny
\setlength{\tabcolsep}{3.2pt}
\renewcommand{\arraystretch}{1.05}
\begin{tabular*}{\columnwidth}{@{\extracolsep{\fill}}llccc@{}}
\toprule
Model & $H$ & $n$ &
\shortstack{Final label / \\ state conn.} &
\shortstack{Final\\match} \\
\midrule
TinyLlama
& 0  & 9 & $0/9$ / $0/9$ & 0.207 \\
TinyLlama
& 3  & 9 & $8/9$ / $2/9$ & 0.978 \\
TinyLlama
& 10 & 9 & $9/9$ / $9/9$ & 1.000 \\
\addlinespace[1pt]

Yi-1.5-6B
& 0  & 9 & $0/9$ / $0/9$ & 0.252 \\
Yi-1.5-6B
& 3  & 9 & $8/9$ / $9/9$ & 0.978 \\
Yi-1.5-6B
& 10 & 9 & $9/9$ / $9/9$ & 1.000 \\
\addlinespace[1pt]

Qwen2.5-7B
& 0  & 9 & $0/9$ / $0/9$ & 0.262 \\
Qwen2.5-7B
& 3  & 9 & $9/9$ / $9/9$ & 1.000 \\
Qwen2.5-7B
& 10 & 9 & $9/9$ / $9/9$ & 1.000 \\
\addlinespace[1pt]

Qwen2.5-14B
& 0  & 9 & $0/9$ / $0/9$ & 0.210 \\
Qwen2.5-14B
& 3  & 9 & $5/9$ / $5/9$ & 0.911 \\
Qwen2.5-14B
& 10 & 9 & $1/9$ / $1/9$ & 0.701 \\
\addlinespace[1pt]

Qwen2.5-32B
& 0  & 9 & $0/9$ / $0/9$ & 0.232 \\
Qwen2.5-32B
& 3  & 9 & $9/9$ / $9/9$ & 1.000 \\
Qwen2.5-32B
& 10 & 9 & $9/9$ / $9/9$ & 1.000 \\
\bottomrule
\end{tabular*}
\end{table}

\paragraph{Partner segregation and participation concentration jointly sustain fragmentation.}
The mixed four-model threshold grid contains 189 trajectories. It reaches final-round behavioral consensus in 21/189 trajectories, stable full behavioral consensus throughout the final 20 rounds in 0/189, and final score-state connectedness in 2/189; mean final match is 0.329. The active-set-matched random control preserves the exact participating agents and pair count in every round while changing only their partners; it reaches 31/189 behavioral-consensus and 15/189 state-connected endpoints, with mean final match 0.385. Thus, assortative partner selection contributes to fragmentation beyond which agents receive fresh evidence.

The pair-count-matched random control preserves the same number of conversations per round but redistributes both participation and partners. It reaches 78/189 behavioral-consensus and 51/189 state-connected endpoints, with mean final match 0.610. The difference is concentrated in retained-memory trajectories. For $H\in\{3,10\}$, threshold, active-set-matched random, and pair-count-matched random routing respectively reach behavioral consensus in 21/126, 31/126, and 78/126 trajectories; score-state connectedness in 2/126, 15/126, and 51/126; and mean final match 0.398, 0.508, and 0.847. The two random controls have the same mean pair budget, 2.792 pairs per round, but redistributing participation lowers the participation Gini, where zero means equal participation, from 0.381 to 0.033. It also lowers the mean longest consecutive period that any agent remains unpaired from 132.7 to 20.4 rounds. Persistent participation concentration is therefore a separate barrier from partner segregation.

Neither random control reaches final behavioral consensus or score-state connectedness in any of the 63 $H=0$ trajectories. Broader and more evenly allocated exposure therefore helps only when social evidence can accumulate. Structured bridge objectives are not uniquely capable of repair, although their repair reliability and state-space cleanliness remain policy dependent. Full participation does not remove similarity-based fragmentation. In the mixed four-model population, full-coverage similarity matching pairs all ten agents in every round, yet none of 27 trajectories ever reaches full behavioral consensus, and none finishes with behavioral consensus or score-state connectedness. The homogeneous Qwen2.5-32B grid gives the same conclusion for the threshold controller itself: all 126 retained-memory threshold trajectories realize five pairs per round and none reaches transient full consensus, whereas every matched non-threshold policy reaches final and stable consensus in all 18 settings at $\Delta=0.25$. Detailed schedule-validation and paired analyses appear in Supplementary Sec.~B.1.1, and the full-participation coverage contrast appears in Supplementary Sec.~B.1.2.

\begin{table}[h]
\caption{Runtime-feedback results in the mixed four-model population. ``PN'' denotes prior-normalized routing; sampled outputs and the reported state-connectedness endpoint remain based on raw restricted scores. Threshold and matched-random rows pool all history horizons, while bridge rows pool retained memory $H\in\{3,10\}$. Every $H=0$ controller family has zero final behavioral consensus and zero final raw score-state connectedness. Full-coverage similarity pairs all ten agents every round using the perfect matching with the smallest total previous-round score-state distance.}
\label{tab:controllers}
\centering
\scriptsize
\setlength{\tabcolsep}{2.5pt}
\renewcommand{\arraystretch}{1.06}
\begin{tabularx}{\columnwidth}{@{}Xcccc@{}}
\toprule
Policy/subset & $n$ & Label cons. & State conn. & Match \\
\midrule
Threshold similarity, raw
& 189 & 21/189 & 2/189 & 0.329 \\

Threshold similarity, PN
& 189 & 0/189 & 0/189 & 0.158 \\
\addlinespace[1pt]

Active-set random, all $H$
& 189 & 31/189 & 15/189 & 0.385 \\

Pair-count random, all $H$
& 189 & 78/189 & 51/189 & 0.610 \\
\addlinespace[1pt]

Full-coverage similarity, all $H$
& 27 & 0/27 & 0/27 & 0.252 \\

State-component, raw retained
& 18 & 14/18 & 13/18 & 0.907 \\

State-component, PN retained
& 18 & 13/18 & 12/18 & 0.912 \\

Label disagreement, retained
& 18 & 14/18 & 10/18 & 0.911 \\
\addlinespace[1pt]

State-distance, raw retained
& 18 & 3/18 & 2/18 & 0.522 \\

State-distance, PN retained
& 18 & 11/18 & 8/18 & 0.781 \\
\bottomrule
\end{tabularx}
\end{table}

\paragraph{Removing default model preferences changes distance-based routing.}
Prior normalization does not rescue similarity-threshold routing. None of the 189 normalized-threshold trajectories ever reaches full behavioral consensus, ends in behavioral consensus, remains aligned throughout the final 20 rounds, or ends with a connected raw or normalized score-state graph.

The raw and normalized representations produce different eligible-edge graphs and therefore different interaction schedules. Among retained-memory settings, normalization lowers the mean number of pairs per round from 2.792 to 1.691, so this comparison does not hold participation fixed. It nevertheless shows that fixed model-family preferences alone do not explain fragmentation.

State-component routing changes little after normalization. State-distance routing improves from 3/18 to 11/18 final behavioral-consensus trajectories, from 2/18 to 8/18 final raw score-state-connected trajectories, and from 0.522 to 0.781 mean final match. Default score differences between model families can therefore change which cross-group pairs fall inside the controller's target distance range.

Behavioral agreement and score-state connectedness need not coincide. In a representative retained-memory mixed trajectory, all ten agents emit \texttt{vel} in the final round, yet the score-state graph retains two components because cross-family restricted-score distances remain larger than within-family distances. This residual score-distribution structure explains why behavioral consensus and score-state connectedness are reported separately throughout.

\begin{figure*}[t]
    \centering
    \includegraphics[width=0.8\textwidth,height=0.50\textheight,keepaspectratio]{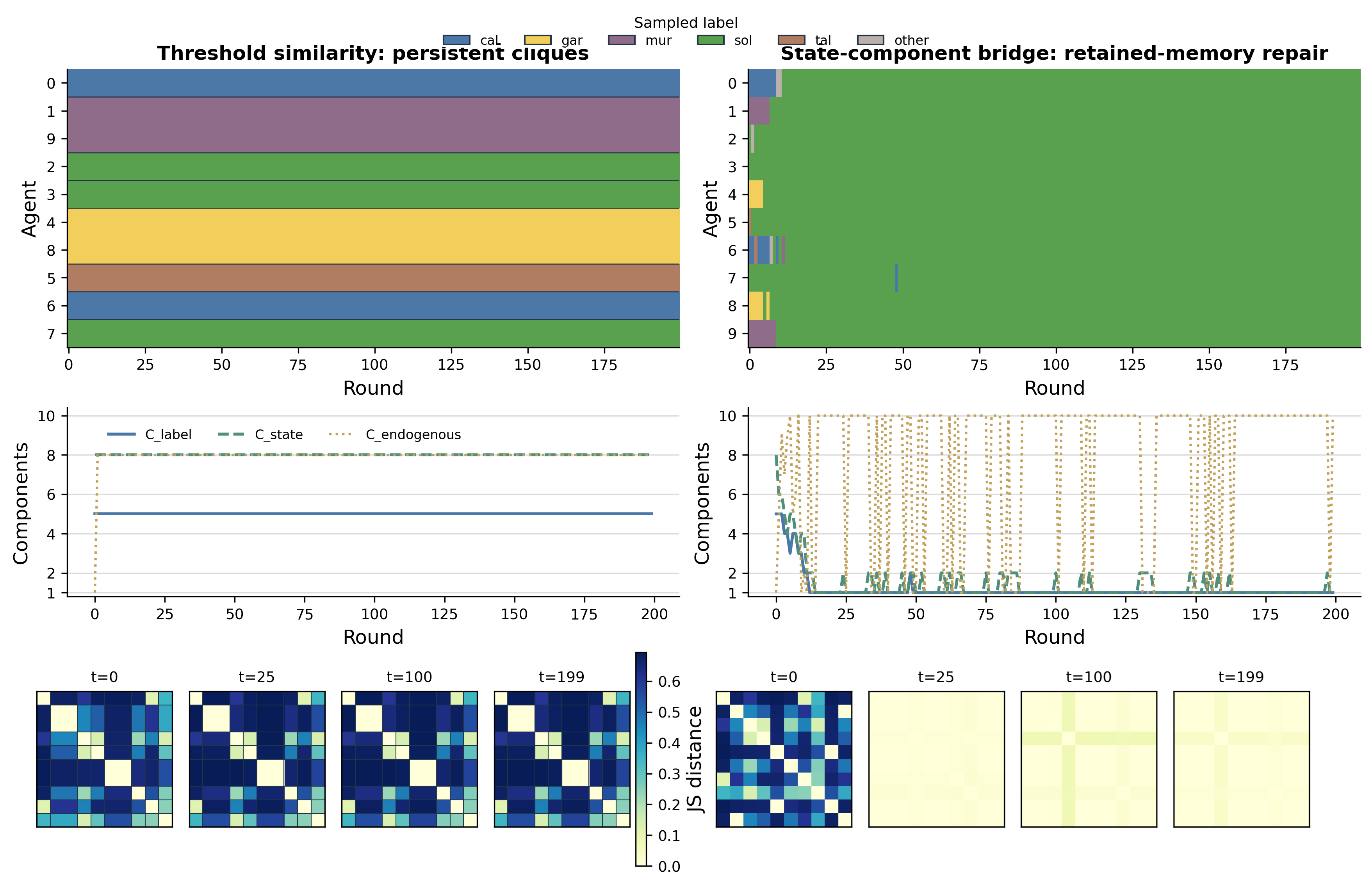}
    \caption{Representative trajectories under similarity-threshold routing (left) and state-component routing with retained history (right). The top panels show each agent's sampled label over time. The middle panels show the numbers of label, score-state, and runtime-candidate-graph components. The bottom panels show pairwise Jensen--Shannon distances at selected rounds. Similarity-threshold routing preserves several persistent groups, whereas state-component routing repeatedly connects different groups and converges in this example.}
    \label{fig:clique_genealogy_main}
\end{figure*}

\paragraph{The effect of homophilous feedback is model conditioned.}
The homogeneous controller grids define the boundary of the central topology contrast across Qwen2.5-14B/32B, Yi-1.5-6B, and TinyLlama-1.1B. Under retained state-component routing, Qwen2.5-32B and Yi reach final behavioral consensus and score-state connectedness in 18/18 settings, TinyLlama reaches them in 17/18 and 12/18, and Qwen2.5-14B reaches both in 7/18 with no stable final-window consensus. Threshold-similarity produces neither final behavioral consensus nor score-state connectedness in any of 189 Qwen2.5-14B or Qwen2.5-32B trajectories and does so in only 1/189 Yi trajectories. TinyLlama supplies the boundary case: retained history permits final behavioral consensus in 45/189 threshold trajectories and score-state connectedness in 35/189. Similarity-threshold routing is therefore not guaranteed to fragment. It fails when groups separate in score space before retained history pulls their preferences close enough for cross-group edges to remain available. Full homogeneous results appear in Supplementary Sec.~B.2.

\paragraph{Consensus attainment does not imply persistence.}
The homogeneous grids also reveal that reaching consensus and maintaining it are distinct response properties. Across 72 matched retained-memory non-threshold settings per model, Qwen2.5-14B and Qwen2.5-32B both attain behavioral consensus in 72/72 trajectories. The 14B populations subsequently exit consensus in 72/72 trajectories, finish in consensus in 31/72, and remain aligned throughout the final 20 rounds in 0/72. Qwen2.5-32B exits after attainment in only 3/72 trajectories and finishes and remains stable in 72/72. The first consensus state is not weaker at 14B: its mean top-label probability is 0.986 versus 0.978 at 32B. The difference therefore arises after attainment rather than from failure to acquire a convention; full trajectory statistics and paired intervals appear in Supplementary Sec.~B.3.

\paragraph{Evidence order contributes to post-attainment disruption.}
To probe this divergence, we held the social records and their label counts fixed and changed only their order. With a 2:1 majority, placing the minority record last makes it top-ranked in 91.6\% of Qwen2.5-14B probes versus 25.0\% at 32B; even against a 4:1 majority, the final minority record overturns the majority in 68.6\% versus 2.7\%. Moving the minority record from the oldest to the newest position increases its log odds relative to the majority by 23.1 and 21.6 at 14B, compared with 9.0 and 5.4 at 32B. A post-consensus population intervention then moves the same conflicting record from the oldest to the newest history position. Newest conflict causes loss of the established target convention in 18/18 runs for each model, whereas oldest conflict does so in 10/18 at 14B and 0/18 at 32B. All intervention runs subsequently recover and remain aligned throughout the final window. Record order therefore causally affects short-term population disruption and may help explain why Qwen2.5-14B repeatedly leaves consensus. Because every intervention run eventually recovers, order sensitivity alone does not explain the persistent 14B behavior. Full prompt-level and routing-resolved results appear in Supplementary Sec.~B.4.

\paragraph{Population composition determines whether exposure is assimilated.}
The retained-memory non-threshold results in Table~\ref{tab:boundary_main} reveal a clear difference in cross-family compatibility. The same-family Qwen population reaches stable behavioral consensus and score-state connectedness under every non-threshold policy. Replacing two 7B agents with TinyLlama preserves stable repair under state-component, label-disagreement, and well-mixed routing, although state-distance fails. Replacing those TinyLlama agents with Yi changes stable repair into transient or absent consensus. This boundary is not explained by homogeneous coordination ability. Under retained state-component bridge routing, homogeneous Yi reaches final behavioral consensus and score-state connectedness in 18/18 settings, whereas homogeneous TinyLlama reaches final behavioral consensus in 17/18 settings and score-state connectedness in 12/18. The mixed-population failure therefore depends on which model families are combined, not on an inability of Yi to coordinate by itself. The Qwen 32B/14B + Yi + TinyLlama cohort and the cohort without 14B agents remain fragmented under every policy. In the no-14B cohort, state-component, label-disagreement, and well-mixed routing repeatedly terminate at the same family-aligned \texttt{sol:8}/\texttt{vel:2} split; state-distance remains more fragmented rather than producing that clean two-basin endpoint.

\begin{table}[h]
\caption{Retained-memory large-cohort outcomes at $H=3$, $\tau=0.3$, $\Delta=0.25$, and three seeds. Each entry lists, in order: final behavioral consensus / final score-state connectedness / stable final-20 behavioral consensus / ever reached full behavioral consensus. Threshold-similarity is 0/3 for all four outcomes in every cohort.}
\label{tab:boundary_main}
\centering
\scriptsize
\setlength{\tabcolsep}{1.0pt}
\renewcommand{\arraystretch}{1.10}
\begin{tabular}{@{}
    >{\raggedright\arraybackslash}p{0.5\columnwidth}
    @{\hspace{1.2pt}}
    cccc
    @{}}
\toprule
Cohort &
\shortstack{State\\comp.} &
\shortstack{Label\\disagr.} &
\shortstack{State\\dist.} &
\shortstack{Well\\mixed} \\
\midrule

Qwen 32B/14B + Yi + TinyLlama
& 0/0/0/0
& 0/0/0/0
& 0/0/0/0
& 0/0/0/0 \\

Qwen2.5 32B/14B/7B
& 3/3/3/3
& 3/3/3/3
& 3/3/3/3
& 3/3/3/3 \\

Qwen 32B/14B/7B + TinyLlama
& 3/3/3/3
& 3/3/3/3
& 0/0/0/0
& 3/3/3/3 \\

Qwen 32B/14B/7B + Yi
& 0/0/0/2
& 0/0/0/2
& 0/0/0/2
& 1/0/0/3 \\

Qwen 32B/7B + Yi + TinyLlama
& 0/0/0/0
& 0/0/0/0
& 0/0/0/0
& 0/0/0/0 \\

\bottomrule
\end{tabular}
\end{table}

\paragraph{Exposure and adoption are distinct.}
We count a directed adoption opportunity when agent $i$ observes agent $j$ while they use different labels. Adoption occurs when $i$ emits $j$'s previous label in the next round:
\begin{align}
P_{\mathrm{adopt}}=\Pr(&y_i(t+1)=y_j(t)\mid i\text{ observed }j,\notag\\
&y_i(t)\neq y_j(t)).
\end{align}
In the no-14B cohort, state-component, label-disagreement, and well-mixed routing each generate roughly one thousand qualifying Yi/non-Yi exposures per direction across three seeds, yet Yi adopts a non-Yi source label in only 1.5--3.0\% of events and non-Yi agents adopt a Yi label in only 0.4--5.7\%. The failure is therefore not explained by a lack of cross-family interaction.

Isolated probes support the same distinction. Yi has no strong default preference for \texttt{vel}, but three consistent history records make the displayed convention top-ranked in only 15/30 \texttt{sol} probes and 22/30 \texttt{vel} probes. All Qwen variants follow either repeated convention in 30/30 probes, whereas TinyLlama follows \texttt{sol} in 30/30 and \texttt{vel} in 0/30. Mixed 2:1 histories also distinguish the homogeneous persistence regimes: the repeated convention remains top-ranked in 87.5\% of Qwen2.5-32B probes but only 7.5\% at 14B. Graph structure controls whether evidence is available; the receiving model controls whether that evidence is assimilated and retained. Full prior, susceptibility, adoption, and seeded-population results are in Supplementary Secs.~C.1--C.2.

\paragraph{Controls and robustness.}
The qualitative ranking of the routing policies persists across state thresholds, vocabulary sizes, and larger populations. At $N\in\{20,50\}$, retained-history Qwen2.5-7B well-mixed populations reach final behavioral consensus and score-state connectedness in 6/6 settings, whereas mixed threshold-similarity reaches neither final nor transient full consensus in any of six settings. State-component and label-disagreement routing still attain transient consensus in 6/6 settings, although terminal repair becomes limited by the fixed $T=200$ budget. History controls separate the effect of partner-label content from the effect of adding more prompt text: in the tested settings, global and within-agent shuffled histories retain full-consensus fractions of $0.835$--$0.995$, cross-agent shuffling lowers them to $0.405$--$0.830$, and removing rendered partner labels yields $0.000$ in every setting. The matched random-routing decomposition quantifies this distinction across the full threshold grid: random rematching produces some repair with the participation schedule fixed and substantially stronger repair when the same pair budget is redistributed across agents. Structured bridge routing is therefore not the only way to obtain coordination. The policies differ in how often agreement is reached, how long it persists, and whether the final score-state graph is connected. Full matched-control and robustness results appear in Supplementary Secs.~B.1.1--B.1.2 and B.5--B.7 and F.

\paragraph{Task-grounded answer choices.}
The task-grounded branch replaces nonce labels with answer choices on ARC-Challenge and MMLU items \cite{hendrycks2021measuring,clark2018think}. Each item run uses $N=6$ homogeneous Qwen2.5-7B agents for $T=20$ rounds. ARC-Challenge uses all 295 four-choice items retained from its 299-item validation set. MMLU-100 contains the 100 highest-entropy questions from a seed-11 pool of 1000 questions drawn from the pooled MMLU validation split. Each task--controller cell aggregates seeds 11--13 and temperatures $\tau\in\{0.3,0.7\}$ with retained history $H=3$ and $\Delta=0.25$. The label-only results reported here use answer labels without rationales or debate; Supplementary Sec.~D.1 reports a full-manifest rationale-sharing contrast under the same four routing policies at seed $11$ and $\tau=0.3$. These experiments are propagation studies rather than benchmark claims: they test how topology changes whether populations settle into correct consensus, wrong consensus, or fragmentation.
Correct consensus means that every agent selects the gold answer. Wrong consensus means that every agent selects the same non-gold answer. Fragmentation means that multiple answer labels remain in the final round.

\begin{table}[h]
\caption{Task-grounded label-only answer-choice validation. ARC-Challenge uses the filtered full validation set with fixed four-choice items ($295$ items); MMLU uses $100$ questions. Rates are over item--seed--temperature runs. $W_{\mathrm{maj}}{\to}C_{\mathrm{maj}}$ counts round-0 wrong strict majority to final correct strict majority; $C_{\mathrm{maj}}{\to}W_{\mathrm{maj}}$ counts the reverse. ``State conn.'' is the rate at which the final restricted
answer-score graph has one connected component.}
\label{tab:task_choice_main}
\centering
\scriptsize
\setlength{\tabcolsep}{2.8pt}
\renewcommand{\arraystretch}{1.05}
\begin{tabular*}{\columnwidth}{@{\extracolsep{\fill}}lcccccc@{}}
\toprule
Policy &
\shortstack{Final\\corr.} &
\shortstack{Correct\\cons.} &
\shortstack{Wrong\\cons.} &
\shortstack{State\\conn.} &
\shortstack{$W_{\mathrm{maj}}{\to}$\\$C_{\mathrm{maj}}$} &
\shortstack{$C_{\mathrm{maj}}{\to}$\\$W_{\mathrm{maj}}$} \\
\midrule
\multicolumn{7}{@{}l}{\textit{ARC-Challenge, $n=1770$ item--seed--temperature runs per policy}}\\
State-comp. bridge & .881 & .834 & .071 & .901 & 30/1770 & 17/1770 \\
Label-disagreement & .878 & .832 & .069 & .894 & 24/1770 & 20/1770 \\
Threshold-sim.     & .870 & .803 & .055 & .848 & 20/1770 & 11/1770 \\
Well-mixed         & .875 & .823 & .074 & .894 & 26/1770 & 29/1770 \\
\addlinespace[1pt]
\multicolumn{7}{@{}l}{\textit{MMLU-100, $n=600$ item--seed--temperature runs per policy}}\\
State-comp. bridge & .336 & .125 & .295 & .567 & 54/600 & 46/600 \\
Label-disagreement & .341 & .145 & .272 & .563 & 55/600 & 44/600 \\
Threshold-sim.     & .333 & .080 & .180 & .330 & 36/600 & 38/600 \\
Well-mixed         & .339 & .138 & .278 & .573 & 54/600 & 49/600 \\
\bottomrule
\end{tabular*}
\end{table}

Table~\ref{tab:task_choice_main} supports the propagation interpretation. On filtered ARC-Challenge validation, all controllers start from a high-correctness regime, so final accuracy differences are modest; nevertheless, threshold-similarity produces less correct consensus and less score-state connectedness than bridge-seeking or well-mixed routing. State-component bridge has the highest final correctness, correct-consensus rate, and score-state- connectedness rate, and it produces more wrong-majority-to-correct-majority flips than the reverse. On MMLU-100, final correctness remains similar across controllers, but topology still changes the collective state: threshold-similarity sharply suppresses both correct and wrong consensus and has much lower score-state connectedness, while bridge and well-mixed policies produce more propagation. Thus the task-choice runs support the graph-control lens without implying that increased propagation is a general accuracy-improvement method.

\section{Discussion}
The results suggest four conditions for coordination. First, agents must remain exposed to other emerging convention groups. Second, interaction opportunities must not remain concentrated on only a few agents. Third, partner-label evidence must stay in the prompt long enough to accumulate. Fourth, the receiving model must respond to and retain that evidence. Even after consensus is reached, subsequent model behavior determines whether it persists.

The matched schedule controls isolate the first two conditions. Full-coverage similarity matching further shows that equal participation is not sufficient when partner selection repeatedly keeps similar agents together. The full-participation Qwen2.5-32B results show that reduced participation is not required for similarity-threshold fragmentation, while the mixed populations show that thresholding can also leave some agents without interactions. Yi-containing populations show that frequent cross-family interaction can fail when adoption is weak, and Qwen2.5-14B shows that consensus can be acquired without being maintained.

The score representation used for routing also matters. Removing each model family's default label preferences leaves state-component routing nearly unchanged but substantially improves state-distance routing. Raw score distances can therefore mix stable differences between model families with changes caused by social interaction. Across the raw-score population contrasts, state-distance routing repairs the same-family mixed-scale Qwen cohort, fails in the Qwen--TinyLlama cohort, and remains weak in the heterogeneous four-model grid. Its effectiveness therefore depends jointly on population geometry and the routing representation rather than being universally effective or ineffective.
Likewise, replacing TinyLlama with Yi while preserving the large-Qwen backbone changes stable repair into transient or absent consensus. Population composition is part of the controlled system, not merely a nuisance variable. Homogeneous coordination ability is therefore not a scalar predictor of heterogeneous compatibility.

The history controls and evidence-order assay show that memory is not a single scalar variable. In the tested 7B and mixed controls, preserving partner-label content while shuffling records within an agent largely preserves consensus, whereas removing partner labels eliminates repair. Partner-label content, rather than prompt length alone, is what brings agents' score states and emitted conventions together. When the retained evidence is conflicting, however, order becomes model conditioned: a terminal minority record disproportionately shifts Qwen2.5-14B and can transiently disrupt an established convention. This distinction reconciles broad robustness to history shuffling with the 14B maintenance regime and shows why increasing the history horizon is not uniformly beneficial.

Agreement in emitted labels and similarity in score distributions are different outcomes. All agents can emit the same label while their score states remain separated, and a connected score-state graph can still contain large pairwise differences. We therefore interpret score-state connectedness jointly with disagreement energy and pairwise Jensen--Shannon distances. The task-grounded contrasts provide a complementary warning: topology can propagate correct or wrong answer basins, so greater consensus is not itself an accuracy guarantee. A practical controller should therefore monitor behavioral agreement, score-state structure, bridge exposure, and task correctness when available, and should continue auditing after first attainment for consensus exits. Persistent fragmentation under limited cross-basin exposure calls for a routing intervention; abundant exposure with low directed adoption instead signals an assimilation or compatibility problem that adding more edges alone is unlikely to solve. The interaction graph is therefore best treated as one element of a closed-loop policy over routing, memory, and cohort composition, rather than as a fixed communication scaffold or a consensus-maximizing objective.

\section{Limitations and Ethics}

The population-composition study evaluates all five routing policies under matched no-memory and retained-memory conditions, $H\in{0,3}$, but fixes $\tau=0.3$, $\Delta=0.25$, $T=200$, and three seeds rather than repeating the full history–temperature–threshold grid for every large cohort. Threshold-similarity can combine assortative partner choice with reduced participation when its candidate graph becomes sparse, although the full-participation controls show that reduced coverage is not required for fragmentation. The matched substitutions therefore establish population-dependent repairability for the tested compositions, not universal properties of Yi, TinyLlama, or Qwen.

The attainment-and-persistence analyses identify behavioral mechanisms rather than complete explanations of model internals. The controlled conflict intervention establishes a causal effect of evidence order on transient disruption, but all runs recover, and the exploratory latent-residue and predictive analyses remain limited in scale and within-grid scope. The matched-random controls use one rematching realization per reference trajectory, while the full-coverage and prior-normalized studies test specific routing and calibration choices rather than all possible alternatives or fixed-exposure counterfactuals.

The task-choice experiments are propagation studies on filtered ARC-Challenge and MMLU-100 with one model cohort, not standard benchmark evaluations. The rationale-sharing contrast uses one seed and temperature, and closed-interface models support only behavioral conclusions because restricted score states are unavailable. No experiment uses personal data or human-subject intervention. Because routing can amplify incorrect as well as correct conventions, consensus should not be interpreted as a safety or quality guarantee.

\section{Conclusion}
Who interacts with whom causally changes collective behavior in language-model populations. Pairing agents only with similar partners can isolate emerging conventions, while interactions across disagreeing groups can support agreement when agents retain partner-label history. The matched controls show that both partner choice and unequal participation matter.
Removing each model family's default label preferences does not eliminate threshold fragmentation, but it improves routing based on intermediate score distances. Even with frequent interaction, some combinations of model families do not adopt one another's conventions, and reaching consensus does not guarantee maintaining it. Agreement in emitted labels can also hide continued disagreement in score distributions, while stronger propagation can spread either correct or incorrect task answers. Collective behavior therefore depends jointly on routing, participation, memory, score representation, population composition, and model response.
The implementation and paper-facing reproduction scripts are available at \url{https://github.com/cedar-lau/llm-graph-control}.


\bibliography{references}
\end{document}


\maketitle
\appendix

\section{Representative regimes and runtime examples}
\label{app:representative_regimes}
\label{app:runtime_threshold_examples}

Table~\ref{tab:representative_app} lists representative terminal regimes from the graph-artifact runs. Table~\ref{tab:runtime_threshold_examples_app} gives representative terminal splits under threshold-similarity routing.
For a round $t$, the dominant-label share and pairwise match rate are
\begin{equation}
    \operatorname{Dom}(t)
    =
    \frac{1}{N}
    \max_{w\in\W}
    \sum_{i=1}^{N}
    \one\{y_i(t)=w\},
    \qquad
    \operatorname{Match}(t)
    =
    \frac{1}{\binom{N}{2}}
    \sum_{i<j}
    \one\{y_i(t)=y_j(t)\}.
    \label{eq:dominance_match_app}
\end{equation}
Thus, $\operatorname{Dom}(t)$ is the fraction of agents using the most common sampled label, while $\operatorname{Match}(t)$ is the fraction of unordered agent pairs whose sampled labels agree. Final behavioral consensus means that the final label graph has one component. Final score-state connectedness means that the thresholded score-state graph has one connected component. Connectedness is a binary topological diagnostic and does not require all pairwise score-state distances to be small; disagreement energy and pairwise Jensen--Shannon distances are reported separately.

\begin{table}[h!]
\caption{Representative final regimes from graph-artifact runs. Components are label/state component counts.}
\label{tab:representative_app}
\centering
\scriptsize
\setlength{\tabcolsep}{4pt}
\begin{tabularx}{\textwidth}{p{0.19\textwidth}p{0.18\textwidth}p{0.24\textwidth}p{0.12\textwidth}Y}
\toprule
Population & Setting & Final labels & Comps. & Regime note \\
\midrule
Qwen2.5-7B & $H=0$, all $\tau$ & \texttt{dek:6, kov:4} & 2/2 & Two-clique split. \\
Qwen2.5-7B & $H=3,10$, all $\tau$ & \texttt{zel:10} & 1/1 & Behavioral consensus and score-state connectedness. \\
Qwen2-7B & $H=0,\tau=0.1,0.3$ & \texttt{dek:6, lod:2, mur:2} & 3/3 & Dominant clique plus fragments. \\
TinyLlama & Representative $H=10$ retained-memory run & \texttt{sol:10} & 1/1 & Behavioral consensus and score-state connectedness. \\
Mixed four-model & $H=3,\tau=0.1$ & \texttt{vel:5, sol:5} & 2/2 & Balanced mixed split. \\
Mixed four-model & $H=3,\tau=0.3$ & \texttt{vel:10} & 1/2 & Behavioral consensus with state residue. \\
Mixed four-model & $H=10,\tau=0.7$ & \texttt{vel:8, sol:2} & 2/2 & Near-consensus with residual clique. \\
\bottomrule
\end{tabularx}
\end{table}

\begin{table}[h]
\caption{Representative runtime threshold-similarity outcomes. ``Dom.'' and ``Match'' are the final-round dominant-label share and pairwise sampled-label agreement rate defined in Eq.~\eqref{eq:dominance_match_app}. $C_{\mathrm{label}}$, $C_{\mathrm{state}}$, and $C_{\mathrm{end}}$ are final label, state-similarity, and endogenous runtime-graph component counts.}
\label{tab:runtime_threshold_examples_app}
\centering
\scriptsize
\setlength{\tabcolsep}{4pt}
\renewcommand{\arraystretch}{1.12}
\begin{tabularx}{\textwidth}{@{}llXccc@{}}
\toprule
Population & Setting & Final labels & Dom. & Match & $C_{\mathrm{label}}/C_{\mathrm{state}}/C_{\mathrm{end}}$ \\
\midrule
Qwen2.5-7B & $H=0,\tau=0.1,\Delta=0.05$ & \texttt{kov:3, bam:2, fel:2, rol:2, cal:1} & 0.30 & 0.133 & 5/10/10 \\
Qwen2.5-7B & $H=0,\tau=0.3,\Delta=0.10$ & \texttt{kov:3, bam:2, cal:2, lod:2, rol:1} & 0.30 & 0.133 & 5/10/10 \\
Qwen2.5-7B & $H=3,\tau=0.3,\Delta=0.20$ & \texttt{kov:6, cal:2, dek:2} & 0.60 & 0.378 & 3/3/3 \\
Mixed four-model & $H=0,\tau=0.1,\Delta=0.15$ & \texttt{sol:5, cal:2, gar:1, lod:1, tal:1} & 0.50 & 0.244 & 5/10/10 \\
Mixed four-model & $H=0,\tau=0.3,\Delta=0.40$ & \texttt{sol:3, cal:1, fel:1, gar:1, lod:1, pel:1, tal:1, vel:1} & 0.30 & 0.0667 & 8/10/10 \\
Mixed four-model & $H=3,\tau=0.3,\Delta=0.20$ & \texttt{sol:3, cal:2, gar:2, mur:2, tal:1} & 0.30 & 0.133 & 5/8/8 \\
\bottomrule
\end{tabularx}
\end{table}

\section{Robustness and mechanism analyses}
\label{app:robustness_checks}
\subsection{Multi-seed robustness}
\label{app:multiseed_robustness}

Table~\ref{tab:multiseed_threshold_mixed_app} summarizes the three-seed mixed four-model threshold-similarity grid. The grid covers $H\in\{0,3,10\}$, $\tau\in\{0.1,0.3,0.7\}$, and seven similarity thresholds, giving 63 configurations and 189 configuration--seed trajectories. Final behavioral consensus occurs in 21/189 trajectories, while final score-state connectedness occurs in 2/189. None of the 189 trajectories remains in full behavioral consensus throughout the final 20 rounds. All final behavioral-consensus and score-state-connected endpoints occur in seed 11; seeds 12 and 13 contribute no final consensus endpoint in 126 trajectories. Threshold routing can therefore reach terminal behavioral agreement, but that agreement is seed dependent, unstable, and only rarely accompanied by a connected score state.

\begin{table}[h]
\caption{Three-seed mixed four-model threshold-similarity results. Slash-separated entries report final behavioral consensus / score-state connectedness and mean final label / score-state component counts.}
\label{tab:multiseed_threshold_mixed_app}
\centering
\scriptsize
\setlength{\tabcolsep}{6pt}
\renewcommand{\arraystretch}{1.12}
\begin{tabularx}{\textwidth}{@{}Xcccc@{}}
\toprule
Subset &
\shortstack{Configs.\\/ trajectories} &
\shortstack{Final\\label / state conn.} &
Final match &
\shortstack{Mean comps.\\label / state} \\
\midrule
All threshold settings
& 63 / 189
& $21/189$ / $2/189$
& 0.329
& 4.20 / 6.86 \\

$H=0$
& 21 / 63
& $0/63$ / $0/63$
& 0.193
& 5.21 / 8.70 \\

$H\in\{3,10\}$
& 42 / 126
& $21/126$ / $2/126$
& 0.398
& 3.70 / 5.94 \\
\bottomrule
\end{tabularx}
\end{table}

Table~\ref{tab:multiseed_bridge_app} expands the main-paper controller results by reporting mean final component counts. No no-memory bridge variant reaches final behavioral consensus or score-state connectedness. With retained memory, state-component and label-disagreement routing reach final behavioral consensus in 14/18 configuration--seed trajectories each, while state-distance routing does so in 3/18. State-component routing also gives the highest score-state-connectedness count in this grid.

\begin{table}[h]
\caption{Three-seed mixed four-model bridge-seeking robustness at $\Delta=0.25$. Each no-memory row contains three configurations over three seeds; each retained-memory row contains six configurations over three seeds. Slash-separated entries report final behavioral consensus / score-state graph connectedness and mean label / score-state component counts.}
\label{tab:multiseed_bridge_app}
\centering
\scriptsize
\setlength{\tabcolsep}{5pt}
\renewcommand{\arraystretch}{1.12}
\begin{tabularx}{\textwidth}{@{}p{0.24\textwidth}lccXc@{}}
\toprule
Runtime policy & Subset & $n$ &
\begin{tabular}[c]{@{}c@{}}Final label /\\state conn.\end{tabular} &
\begin{tabular}[c]{@{}c@{}}Mean comps.\\label/state\end{tabular} \\
\midrule
State-component bridge & $H=0$ & 9
& $0/9$ / $0/9$
& 4.78 / 7.22 \\
State-component bridge & $H\in\{3,10\}$ & 18
& $14/18$ / $13/18$
& 1.22 / 1.44 \\
Label disagreement & $H=0$ & 9
& $0/9$ / $0/9$
& 4.89 / 6.89 \\
Label disagreement & $H\in\{3,10\}$ & 18
& $14/18$ / $10/18$
& 1.22 / 1.61 \\
State-distance bridge & $H=0$ & 9
& $0/9$ / $0/9$
& 5.33 / 7.22 \\
State-distance bridge & $H\in\{3,10\}$ & 18
& $3/18$ / $2/18$
& 2.17 / 2.44 \\
\bottomrule
\end{tabularx}
\end{table}

A matched homogeneous Qwen2.5-7B contrast gives the same qualitative separation. Threshold-similarity at $H=3,\tau=0.3,\Delta=0.20$ remains fragmented in all three seeds, with final label/state component counts $3/3$, $4/4$, and $3/3$. State-component routing at the same $H$ and $\tau$ reaches final behavioral consensus and score-state connectedness in all three seeds. The homogeneous and heterogeneous contrasts therefore agree that similarity-dependent routing can preserve separated basins, whereas retained cross-component exposure can support convergence.

\subsubsection{Matched schedule decomposition}
\label{app:matched_schedule_decomposition}

Threshold-similarity jointly determines which agents participate and which participating agents meet. We separate these two effects by replaying every one of the 189 threshold schedules under two randomized routing policies. Let $A_t$ denote the agents appearing in the realized threshold matching at round $t$, and let $k_t=|A_t|/2$ be the realized pair count. The active-set-matched random policy uniformly rematches the agents in $A_t$, preserving the exact active-agent identities, unpaired agents, pair count, per-agent participation totals, and starvation schedule while changing partner identities. The pair-count-matched random policy preserves only $k_t$: it randomly selects $2k_t$ agents and randomly pairs them, holding the interaction budget fixed while allowing participation opportunities to move across the population.

Both policy families inherit the corresponding threshold trajectory's resolved model ordering, prompts, tokenizer-safe labels, decoding settings, initialization, and reward bookkeeping. Their schedules are generated before inference and do not depend on the labels or score states subsequently produced by the control trajectory. All 189 active-set-matched and all 189 pair-count-matched trajectories complete 200 rounds and pass their schedule-replay audits.

\begin{table}[h]
\caption{Matched schedule decomposition in the mixed four-model population. Active-set random preserves the exact threshold participants in every round; pair-count random preserves only the number of pairs. Slash-separated entries report final behavioral consensus / score-state connectedness and mean final label / score-state component counts.}
\label{tab:matched_schedule_endpoints_app}
\centering
\scriptsize
\setlength{\tabcolsep}{5pt}
\renewcommand{\arraystretch}{1.10}
\begin{tabularx}{\textwidth}{@{}Xlcccc@{}}
\toprule
Policy & Subset & $n$ &
\shortstack{Final\\label / state} &
Final match &
\shortstack{Mean comps.\\label / state} \\
\midrule
Threshold similarity
& all $H$ & 189
& $21/189$ / $2/189$
& 0.329
& 4.20 / 6.86 \\

Active-set random
& all $H$ & 189
& $31/189$ / $15/189$
& 0.385
& 4.05 / 5.85 \\

Pair-count random
& all $H$ & 189
& $78/189$ / $51/189$
& 0.610
& 2.93 / 4.32 \\
\addlinespace[2pt]

Threshold similarity
& retained & 126
& $21/126$ / $2/126$
& 0.398
& 3.70 / 5.94 \\

Active-set random
& retained & 126
& $31/126$ / $15/126$
& 0.508
& 3.28 / 4.80 \\

Pair-count random
& retained & 126
& $78/126$ / $51/126$
& 0.847
& 1.54 / 2.40 \\
\bottomrule
\end{tabularx}
\end{table}

All 63 no-memory trajectories in each policy family finish without behavioral consensus or score-state connectedness. The retained-memory contrasts therefore isolate how routing changes the propagation of evidence that can accumulate across rounds.

Because active-set random preserves the complete participation schedule, its improvement over threshold routing isolates partner selection under the same interaction coverage. Behavioral-consensus incidence rises by 0.079, score-state-connectedness incidence by 0.103, and mean final match by 0.110. Pair-count random produces a larger change while preserving the same number of conversations: relative to threshold routing, behavioral-consensus incidence rises by 0.452, score-state connectedness by 0.389, and mean final match by 0.450. Table~\ref{tab:matched_schedule_effects_app} reports paired uncertainty estimates.

\begin{table}[h]
\caption{Paired effects for the 126 retained-memory trajectories. Entries are rate or mean differences with 95\% percentile intervals from 20,000 bootstrap resamples of the 42 $(H,\tau,\Delta)$ configurations, retaining all three seeds within each sampled configuration. Positive values favor the policy named first.}
\label{tab:matched_schedule_effects_app}
\centering
\scriptsize
\setlength{\tabcolsep}{5pt}
\renewcommand{\arraystretch}{1.12}
\begin{tabularx}{\textwidth}{@{}Xccc@{}}
\toprule
Contrast &
$\Delta$ label consensus &
$\Delta$ state connectedness &
$\Delta$ final match \\
\midrule
Active-set random $-$ threshold
& $+0.079\,[0.008,0.151]$
& $+0.103\,[0.056,0.151]$
& $+0.110\,[0.073,0.148]$ \\

Pair-count random $-$ threshold
& $+0.452\,[0.373,0.532]$
& $+0.389\,[0.294,0.492]$
& $+0.450\,[0.410,0.490]$ \\

Pair-count random $-$ active-set random
& $+0.373\,[0.294,0.452]$
& $+0.286\,[0.198,0.381]$
& $+0.339\,[0.298,0.380]$ \\
\bottomrule
\end{tabularx}
\end{table}

The difference between the two randomized policies arises from participation allocation rather than interaction quantity. Their retained-memory pair budgets and paired-agent fractions are identical, but pair-count random spreads those opportunities much more evenly and produces more persistent agreement.

\begin{table}[h]
\caption{Participation and temporal outcomes for the retained-memory randomized policies. ``Stable/trans.'' reports stable final-20 / transient full behavioral-consensus counts, and ``Full frac.'' is the mean fraction of rounds with full behavioral consensus.}
\label{tab:matched_schedule_participation_app}
\centering
\scriptsize
\setlength{\tabcolsep}{4pt}
\renewcommand{\arraystretch}{1.10}
\begin{tabularx}{\textwidth}{@{}Xcccccc@{}}
\toprule
Policy &
\shortstack{Mean\\pairs} &
\shortstack{Paired\\fraction} &
\shortstack{Participation\\Gini} &
\shortstack{Mean max.\\unpaired streak} &
Stable/trans. &
Full frac. \\
\midrule
Active-set random
& 2.792 & 0.558 & 0.381 & 132.7
& $28/126$ / $37/126$
& 0.203 \\

Pair-count random
& 2.792 & 0.558 & 0.033 & 20.4
& $63/126$ / $102/126$
& 0.461 \\
\bottomrule
\end{tabularx}
\end{table}

At the configuration-cluster level, pair-count random lowers participation Gini relative to active-set random by 0.348, with a 95\% interval of $[0.324,0.373]$, and lowers the maximum unpaired streak by 112.3 rounds, with an interval of $[104.9,118.6]$. Its stable-consensus rate is higher by 0.278 $[0.198,0.357]$, and its transient-consensus rate is higher by 0.516 $[0.452,0.579]$.

The allocation effect is clearest in the sparse seed-12 and seed-13 retained-memory schedules. Active-set random preserves their participation patterns and reaches 0/84 final behavioral-consensus and 0/84 state-connected endpoints. Pair-count random uses the same mean budget of 1.688 pairs per round but reaches 52/84 behavioral-consensus and 36/84 state-connected endpoints. Mean participation Gini falls from 0.571 to 0.049, and the mean maximum unpaired streak falls from 199.0 to 30.5 rounds. Seed 11 already supplies five pairs per round with equal participation, so its differences between the two randomized policies reflect alternative rematching realizations rather than participation redistribution.

These contrasts separate two mechanisms within the sparse threshold schedules. Similarity-based partner selection contributes even when the active-agent identities and unpaired periods are held fixed, while repeatedly concentrating a fixed interaction budget on the same agents creates an additional and larger barrier. Supplementary Sec.~B.1.2 removes participation loss entirely: under full-coverage minimum-distance matching, all ten agents interact in every round, yet
none of 27 trajectories reaches full behavioral consensus. Thus, under
the tested full-coverage similarity rule, partner choice can sustain
fragmentation even with equal participation, while uneven participation
adds a separate barrier in sparse threshold regimes.

\subsubsection{Full-coverage similarity and sparse-regime contrasts}
\label{app:interaction_coverage}

The full-coverage similarity condition tests whether a preference for similar partners can preserve fragmentation when participation loss is removed inside the mixed four-model population. All ten agents participate in every round. At round $0$, the controller forms a complete matching before score states are available. For each later round, it selects the perfect matching that minimizes the total previous-round Jensen--Shannon distance across the five selected pairs. Equivalently, it computes a maximum-weight perfect matching with edge weight $-D_{\mathrm{JS}}(s_i(t-1),s_j(t-1))$.

The controller uses no similarity cutoff and no fallback. All 5400 saved rounds contain five pairs and no unpaired agent, and each scored matching attains the minimum possible total partner distance. Participation Gini and maximum unpaired streak are therefore zero in every trajectory. The grid covers $H\in\{0,3,10\}$, $\tau\in\{0.1,0.3,0.7\}$, and seeds 11--13, giving 27 trajectories of 200 rounds each.

\begin{table}[h]
\caption{Full-coverage similarity matching in the mixed four-model population. Every round contains five pairs and no unpaired agent. ``Final label / state'' reports final behavioral consensus and final score-state connectedness. ``Stable/trans.'' reports stable final-20 and ever-reached full behavioral consensus. ``Pair JS'' is the mean previous-round Jensen--Shannon distance among selected partners.}
\label{tab:full_coverage_similarity_app}
\centering
\scriptsize
\setlength{\tabcolsep}{4pt}
\renewcommand{\arraystretch}{1.12}
\begin{tabularx}{\textwidth}{@{}Xcccccc@{}}
\toprule
Subset & $n$ &
\shortstack{Final label /\\state conn.} &
Stable/trans. &
Final match &
\shortstack{Mean comps.\\label/state} &
Pair JS \\
\midrule
All $H$
& 27 & $0/27$ / $0/27$
& $0/27$ / $0/27$
& 0.252 & 3.81 / 4.00 & 0.042 \\

$H=0$
& 9 & $0/9$ / $0/9$
& $0/9$ / $0/9$
& 0.188 & 4.67 / 5.00 & 0.074 \\

$H=3$
& 9 & $0/9$ / $0/9$
& $0/9$ / $0/9$
& 0.279 & 3.44 / 3.67 & 0.019 \\

$H=10$
& 9 & $0/9$ / $0/9$
& $0/9$ / $0/9$
& 0.289 & 3.33 / 3.33 & 0.031 \\
\bottomrule
\end{tabularx}
\end{table}

Full participation does not produce global coordination under this similarity objective. None of the 27 trajectories reaches full behavioral consensus in any round, and none finishes with behavioral consensus or a connected score-state graph. Retained history nevertheless improves partial alignment: mean final match rises from 0.188 at $H=0$ to 0.284 across $H\in\{3,10\}$, while mean final label/state component counts fall from $4.67/5.00$ to $3.39/3.50$. Partner-label history therefore brings agents closer within the existing groups without merging those groups when partner selection continues to favor the most similar available agents.

The contrast with full-coverage bridge routing isolates the importance of the partner objective. Across the same 18 retained-memory history--temperature--seed settings, full-coverage similarity reaches $0/18$ final or transient behavioral consensus and has mean final match 0.284. State-component and label-disagreement routing each reach final behavioral consensus in $14/18$ settings, with mean final matches 0.907 and 0.911, while state-distance routing reaches $3/18$ with mean match 0.522. All of these controllers provide every agent with an interaction; what differs is whether the selected pairs preserve or cross the emerging score-state groups.

Together with the matched schedule controls, this completes the partner-choice and participation decomposition. Random partner reassignment improves coordination under the threshold participation schedule, redistributing the same interaction budget improves it further, and removing participation loss entirely does not restore coordination when the matching continues to favor similar partners. The homogeneous Qwen2.5-32B grid provides a complementary full-participation contrast using the threshold controller itself, while the larger heterogeneous cohorts show how thresholding can additionally reduce participation. Table~\ref{tab:coverage_audit_app} summarizes these full- and partial-participation regimes.

\begin{table}[h]
\caption{Realized interaction coverage under threshold and non-threshold routing. ''Pair frac.'' is the mean fraction of agents paired in a round. ''Max unpaired'' is zero when every agent is always paired; for the large-cohort aggregate it reports the number of trajectories containing a 199-round unpaired streak. All rows use $N=10$. The final column gives transient / final behavioral consensus.}
\label{tab:coverage_audit_app}
\centering
\scriptsize
\setlength{\tabcolsep}{4pt}
\renewcommand{\arraystretch}{1.10}
\begin{tabularx}{\textwidth}{@{}Xccccc@{}}
\toprule
Setting & $n$ &
\shortstack{Mean\\pairs} &
\shortstack{Pair\\frac.} &
\shortstack{Max\\unpaired} &
\shortstack{Transient /\\final cons.} \\
\midrule
Mixed full-coverage similarity, all $H$
& 27 & 5.00 & 1.000 & 0 rounds & 0/27 / 0/27 \\

32B threshold, retained, all $\Delta$
& 126 & 5.00 & 1.000 & 0 rounds & 0/126 / 0/126 \\

32B threshold, retained, $\Delta=0.25$
& 18 & 5.00 & 1.000 & 0 rounds & 0/18 / 0/18 \\

32B well mixed, retained
& 18 & 5.00 & 1.000 & 0 rounds & 18/18 / 18/18 \\

32B state-component, retained
& 18 & 5.00 & 1.000 & 0 rounds & 18/18 / 18/18 \\

32B label disagreement, retained
& 18 & 5.00 & 1.000 & 0 rounds & 18/18 / 18/18 \\

32B state-distance, retained
& 18 & 5.00 & 1.000 & 0 rounds & 18/18 / 18/18 \\

Large cohorts, threshold, $H=3$
& 15 & 2.80 & 0.559 & 14/15 at 199 rounds & 0/15 / 0/15 \\
\bottomrule
\end{tabularx}
\end{table}

The retained-memory Qwen2.5-32B comparison isolates partner selection at a fixed interaction budget. Threshold routing and every comparator realize five pairs per round, yet none of the 126 threshold trajectories reaches transient, final, or stable behavioral consensus. At $\Delta=0.25$, each matched non-threshold policy instead reaches final and stable consensus in 18/18 settings. Threshold fragmentation therefore does not require a reduction in the number of interactions.

The large heterogeneous cohorts reveal a second consequence of the same feedback rule. Across the 15 retained-memory threshold trajectories, mean coverage falls to 2.80 of five possible pairs per round, and 14/15 trajectories contain an agent that remains unmatched throughout all 199 post-initial rounds. In these populations, threshold feedback both restricts interactions to score-similar partners and produces uneven access to fresh social evidence. Full-participation segregation and sparse-graph participation loss are therefore complementary population-dependent regimes rather than competing explanations.

\subsubsection{Score-state connectedness and concentration}
\label{app:connectedness_concentration}

Graph connectedness is weaker than complete score-state contraction. To quantify the distinction, we compute all pairwise Jensen--Shannon distances between agents' restricted score distributions using a common diagnostic threshold $\Delta=0.25$. Across the analyzed trajectories, 12.4\% of rounds with full behavioral consensus retain a disconnected score-state graph. Among connected score-state graphs, 12.0\% have maximum pairwise distance above 0.20, 5.8\% exceed 0.25, and 2.1\% exceed 0.35.

These frequencies aggregate several controller and population regimes and are therefore descriptive rather than estimates of one common treatment effect. They demonstrate that connectedness and concentration capture different properties of the population state, motivating the joint reporting of component counts, disagreement energy, and pairwise-distance summaries.

\subsubsection{Prior-normalized routing geometry}
\label{app:prior_normalized_routing}

Raw restricted-score geometry in a heterogeneous population can combine two sources of variation: stationary model-specific calibration under the shared prompt and interaction-conditioned displacement produced by accumulated social evidence. We separate these sources by estimating one prompt-conditioned prior over the common ten-label vocabulary for each of the four model families.

For model family $m$, the prior campaign uses 90 empty-history probes: 30 independently randomized allowed-label orders for each of seeds 11--13. Every score vector is first mapped back to the canonical label order. If $p^{(0)}_{m,r}(w)$ is the restricted probability of label $w$ in probe $r$, the saved prior is the normalized geometric mean
\begin{equation}
    \ell_m(w)
    =
    \frac{1}{90}
    \sum_{r=1}^{90}
    \log\!\left(p^{(0)}_{m,r}(w)+\epsilon\right),
    \qquad
    \pi_m
    =
    \softmax(\ell_m),
    \qquad
    \epsilon=10^{-12}.
    \label{eq:model_prior_normalization_app}
\end{equation}
For agent $i$ of model family $m(i)$, the online prior-normalized routing state is
\begin{equation}
    \widetilde{s}_i(t)
    =
    \softmax\!\left(
        \log(s_i(t)+\epsilon)
        -
        \log\pi_{m(i)}
    \right).
    \label{eq:online_prior_normalized_state_app}
\end{equation}
Only the routing representation is changed. Sampled outputs continue to use the original restricted logits, raw score states are retained for comparison with the rest of the paper, and the controller constructs round-$t+1$ pairings only from normalized states saved at round $t$.

The campaign contains 360 prior probes and 243 population trajectories. Prior-normalized threshold-similarity repeats the full
\[
H\in\{0,3,10\},
\qquad
\tau\in\{0.1,0.3,0.7\},
\qquad
\Delta\in\{0.05,0.10,0.15,0.20,0.25,0.30,0.40\}
\]
grid over seeds 11--13, giving 189 trajectories. Prior-normalized state-component and state-distance routing each use $\Delta=0.25$ over the same history, temperature, and seed grid, giving 27 trajectories per policy. All 243 trajectories complete 200 rounds and retain the resolved model ordering, labels, prompts, decoding settings, initialization, and reward bookkeeping of their matched configurations.

\begin{table}[h]
\caption{Prior-normalized routing outcomes in the mixed four-model population. ``Stable/trans.'' gives stable final-20 / transient full behavioral-consensus counts. ``State conn.'' gives final raw / prior-normalized score-state connectedness at the common diagnostic threshold $\Delta=0.25$. Mean pairs is the average realized number of interactions across all 200 rounds.}
\label{tab:prior_normalized_results_app}
\centering
\tiny
\setlength{\tabcolsep}{3.5pt}
\renewcommand{\arraystretch}{1.12}
\begin{tabularx}{\textwidth}{@{}Xcccccc@{}}
\toprule
Policy/subset &
$n$ &
\shortstack{Final\\behavioral} &
Stable/trans. &
\shortstack{State conn.\\raw / normalized} &
\shortstack{Final\\match} &
\shortstack{Mean\\pairs} \\
\midrule
Threshold similarity, all $H$
& 189
& $0/189$
& $0/189$ / $0/189$
& $0/189$ / $0/189$
& 0.158
& 1.334 \\

Threshold similarity, retained
& 126
& $0/126$
& $0/126$ / $0/126$
& $0/126$ / $0/126$
& 0.173
& 1.691 \\
\addlinespace[2pt]

State-component, $H=0$
& 9
& $0/9$
& $0/9$ / $0/9$
& $0/9$ / $0/9$
& 0.198
& 5.000 \\

State-component, $H=3$
& 9
& $7/9$
& $3/9$ / $9/9$
& $6/9$ / $4/9$
& 0.956
& 5.000 \\

State-component, $H=10$
& 9
& $6/9$
& $6/9$ / $8/9$
& $6/9$ / $0/9$
& 0.869
& 5.000 \\
\addlinespace[2pt]

State-distance, $H=0$
& 9
& $0/9$
& $0/9$ / $0/9$
& $0/9$ / $0/9$
& 0.170
& 5.000 \\

State-distance, $H=3$
& 9
& $8/9$
& $6/9$ / $9/9$
& $5/9$ / $3/9$
& 0.978
& 5.000 \\

State-distance, $H=10$
& 9
& $3/9$
& $2/9$ / $3/9$
& $3/9$ / $0/9$
& 0.585
& 5.000 \\
\bottomrule
\end{tabularx}
\end{table}

Table~\ref{tab:prior_normalized_comparison_app} places the primary retained-memory endpoints beside the corresponding raw-score routing results. For comparability with the rest of the paper, the state-connectedness entry in this table is evaluated on the raw restricted-score state even when normalized states determine the routing graph.

\begin{table}[h]
\caption{Raw-score and prior-normalized routing in retained-memory mixed populations. Each entry reports final behavioral consensus / final raw score-state connectedness / mean final match. Threshold rows contain 126 trajectories; bridge rows contain 18 trajectories.}
\label{tab:prior_normalized_comparison_app}
\centering
\scriptsize
\setlength{\tabcolsep}{6pt}
\renewcommand{\arraystretch}{1.12}
\begin{tabularx}{\textwidth}{@{}Xcc@{}}
\toprule
Policy &
Raw-score routing &
Prior-normalized routing \\
\midrule
Threshold similarity
& $21/126$ / $2/126$ / 0.398
& $0/126$ / $0/126$ / 0.173 \\

State-component
& $14/18$ / $13/18$ / 0.907
& $13/18$ / $12/18$ / 0.912 \\

State-distance
& $3/18$ / $2/18$ / 0.522
& $11/18$ / $8/18$ / 0.781 \\
\bottomrule
\end{tabularx}
\end{table}

The raw threshold grid reaches final-round behavioral consensus in 21/189 trajectories, stable behavioral consensus throughout the final 20 rounds in 0/189, and final score-state connectedness in 2/189. After prior normalization, none of the 189 threshold trajectories reaches even transient full behavioral consensus, and none finishes with behavioral consensus or with a connected raw or normalized score-state graph. The two threshold grids are not fixed-schedule counterfactuals: normalization changes the candidate graph and lowers mean retained-memory coverage from 2.792 to 1.691 pairs per round. The result therefore does not isolate socially induced geometry under identical exposure. It does show that stationary model calibration is not a sufficient explanation for threshold fragmentation, because removing that calibration does not restore a viable coordination regime anywhere in the tested sweep.

State-component routing is comparatively insensitive to the change of representation. With retained memory, final behavioral consensus changes from 14/18 to 13/18, final raw-state connectedness from 13/18 to 12/18, and mean final match from 0.907 to 0.912. The component objective therefore continues to create useful cross-basin exposure when its components are formed from model-relative rather than raw score states.

State-distance routing changes much more strongly. Prior normalization raises retained-memory final behavioral consensus from 3/18 to 11/18, final raw-state connectedness from 2/18 to 8/18, and mean final match from 0.522 to 0.781. The gain is concentrated at $H=3$, where 8/9 trajectories finish in behavioral consensus, 6/9 satisfy stable final-20 consensus, and mean final match reaches 0.978. At $H=10$, the corresponding outcomes are 3/9 final, 2/9 stable, and 0.585 mean match. Longer retained history therefore remains model- and controller-conditioned rather than being monotonically beneficial.

Raw and normalized connectedness answer different questions. Raw connectedness asks whether the models' actual restricted output distributions form one graph in the common score coordinates used throughout the paper. Normalized connectedness asks whether their deviations relative to model-specific stationary baselines form one graph. Among retained-memory runs, prior-normalized state-component routing reaches raw connectedness in 12/18 trajectories but normalized connectedness in 4/18; state-distance reaches the two endpoints in 8/18 and 3/18. To use a common endpoint across routing representations, the primary comparisons report raw-state connectedness, while normalized-state connectedness records residual structure in model-relative coordinates.

Together, the prior-normalized results distinguish two roles of stationary calibration. It is not sufficient to explain the fragmentation produced by threshold-similarity, but it can distort the ranking of candidate edges for a metric bridge objective. Calibration-aware state representations can therefore improve which cross-basin interactions are selected without making homophilous routing intrinsically safe.

\subsection{Homogeneous model-family and scale contrasts}
\label{app:model_scale_robustness}

The homogeneous suite evaluates TinyLlama-1.1B-Chat, Yi-1.5-6B-Chat, and Qwen2.5-Instruct populations at 7B, 14B, and 32B under a shared $N=10$, $M=10$, and $T=200$ protocol. The well-mixed history sweeps use seeds $11$--$13$, temperatures
$\tau\in\{0.1,0.3,0.7\}$, and history horizons $H\in\{0,3,10\}$. TinyLlama, Yi, Qwen2.5-14B, and Qwen2.5-32B also use complete controller grids. For each of those four models, threshold-similarity is evaluated over
\[
    \Delta\in\{0.05,0.10,0.15,0.20,0.25,0.30,0.35\},
\]
giving 189 threshold setting--seed rows, while state-component, label-disagreement, and state-distance bridge routing each contribute 27 rows at $\Delta=0.25$. Together with the 27 well-mixed rows, this gives 297 trajectories per complete model grid and 1188 trajectories across the four grids. Qwen2.5-7B contributes the matched well-mixed history sweep and the targeted controller contrasts reported in Supplementary Sections~B.1, B.5--B.7, and F.

\begin{table}[h]
\caption{Homogeneous well-mixed history sweep across five open-weight models. Rows pool seeds $11$--$13$ and temperatures $\tau\in\{0.1,0.3,0.7\}$ within each history horizon. ``Final label / state conn.'' reports final behavioral-consensus and score-state-connectedness counts.}
\label{tab:homogeneous_history_all_models_app}
\centering
\scriptsize
\setlength{\tabcolsep}{4.5pt}
\renewcommand{\arraystretch}{1.08}
\begin{tabular}{@{}lccccc@{}}
\toprule
Model & $H$ & $n$ &
\shortstack{Final label /\\state conn.} &
\shortstack{Final\\match} \\
\midrule
TinyLlama-1.1B & 0  & 9 & $0/9$ / $0/9$ & 0.207 \\
TinyLlama-1.1B & 3  & 9 & $8/9$ / $2/9$ & 0.978 \\
TinyLlama-1.1B & 10 & 9 & $9/9$ / $9/9$ & 1.000 \\
\addlinespace[1pt]

Yi-1.5-6B & 0  & 9 & $0/9$ / $0/9$ & 0.252 \\
Yi-1.5-6B & 3  & 9 & $8/9$ / $9/9$ & 0.978 \\
Yi-1.5-6B & 10 & 9 & $9/9$ / $9/9$ & 1.000 \\
\addlinespace[1pt]

Qwen2.5-7B & 0  & 9 & $0/9$ / $0/9$ & 0.262 \\
Qwen2.5-7B & 3  & 9 & $9/9$ / $9/9$ & 1.000 \\
Qwen2.5-7B & 10 & 9 & $9/9$ / $9/9$ & 1.000 \\
\addlinespace[1pt]

Qwen2.5-14B & 0  & 9 & $0/9$ / $0/9$ & 0.210 \\
Qwen2.5-14B & 3  & 9 & $5/9$ / $5/9$ & 0.911 \\
Qwen2.5-14B & 10 & 9 & $1/9$ / $1/9$ & 0.701 \\
\addlinespace[1pt]

Qwen2.5-32B & 0  & 9 & $0/9$ / $0/9$ & 0.232 \\
Qwen2.5-32B & 3  & 9 & $9/9$ / $9/9$ & 1.000 \\
Qwen2.5-32B & 10 & 9 & $9/9$ / $9/9$ & 1.000 \\
\bottomrule
\end{tabular}
\end{table}

Table~\ref{tab:homogeneous_history_all_models_app} shows that retained social evidence changes the homogeneous regime across all five models. None of the 45 $H=0$ runs reaches final behavioral consensus or score-state connectedness, whereas every model reaches transient full behavioral consensus in all 18 retained-history runs. Terminal behavior remains model conditioned. TinyLlama at $H=3$ reaches final behavioral consensus in 8/9 runs but score-state connectedness in only 2/9, revealing substantial latent residue; increasing the horizon to $H=10$ yields both forms of consensus in 9/9. Yi reaches score-state connectedness in all 18 retained-history runs and final behavioral consensus in 17/18. Qwen2.5-14B is the principal persistence exception, while Qwen2.5-7B and Qwen2.5-32B repair completely.

The Qwen2.5-14B/32B comparison below retains the stability and component-level statistics used for the matched attainment-and-persistence analysis.

\begin{table}[h]
\caption{Homogeneous Qwen2.5-14B and Qwen2.5-32B well-mixed history-sweep results. Rows pool seeds $11$--$13$ and temperatures $\tau\in\{0.1,0.3,0.7\}$ within each history horizon. ``Final label / state conn.'' reports final behavioral consensus and score-state connectedness, and ``Stable/trans.'' reports stable and transient full behavioral consensus.}
\label{tab:model_scale_history_app}
\centering
\scriptsize
\setlength{\tabcolsep}{5pt}
\renewcommand{\arraystretch}{1.12}
\begin{tabular}{@{}lcccccc@{}}
\toprule
Model & $H$ & $n$ &
\begin{tabular}[c]{@{}c@{}}Final label /\\state conn.\end{tabular} &
\begin{tabular}[c]{@{}c@{}}Stable/\\trans.\end{tabular} &
\begin{tabular}[c]{@{}c@{}}Final\\match\end{tabular} &
\begin{tabular}[c]{@{}c@{}}Mean comps.\\label/state\end{tabular} \\
\midrule
Qwen2.5-14B & 0  & 9 & $0/9$ / $0/9$ & $0/9$ / $2/9$ & 0.210 & 4.78 / 4.44 \\
Qwen2.5-14B & 3  & 9 & $5/9$ / $5/9$ & $0/9$ / $9/9$ & 0.911 & 1.44 / 1.44 \\
Qwen2.5-14B & 10 & 9 & $1/9$ / $1/9$ & $0/9$ / $9/9$ & 0.701 & 2.11 / 2.22 \\
\addlinespace[2pt]
Qwen2.5-32B & 0  & 9 & $0/9$ / $0/9$ & $0/9$ / $0/9$ & 0.232 & 4.00 / 4.22 \\
Qwen2.5-32B & 3  & 9 & $9/9$ / $9/9$ & $9/9$ / $9/9$ & 1.000 & 1.00 / 1.00 \\
Qwen2.5-32B & 10 & 9 & $9/9$ / $9/9$ & $9/9$ / $9/9$ & 1.000 & 1.00 / 1.00 \\
\bottomrule
\end{tabular}
\end{table}

Table~\ref{tab:model_scale_history_app} separates the incidence of repair from its terminal stability. Qwen2.5-32B gives a complete memory transition: every $H=0$ run remains fragmented, while every $H=3$ and $H=10$ run reaches final behavioral consensus and score-state connectedness and remains behaviorally aligned throughout the final 20 rounds. Qwen2.5-14B shows the same directional dependence on retained evidence but a noisier terminal regime. All retained-history runs reach full behavioral consensus at some point, and $H=3$ produces high final match and final consensus in $5/9$ runs, but none satisfies the final-window stability criterion. The lower terminal rate at $H=10$ further shows that longer retained history is not monotonically beneficial in this model.

\begin{table}[h]
\caption{Homogeneous Qwen2.5-14B and Qwen2.5-32B runtime-controller results. Threshold rows aggregate all $H$, $\tau$, seven $\Delta$ values, and seeds $11$--$13$. Bridge rows use $\Delta=0.25$; ``retained'' pools $H\in\{3,10\}$. ``Final label / state conn.'' reports final behavioral consensus / score-state connectedness, and ``Stable/trans.'' reports stable and transient full behavioral consensus.}
\label{tab:model_scale_controller_app}
\centering
\scriptsize
\setlength{\tabcolsep}{4pt}
\renewcommand{\arraystretch}{1.10}
\begin{tabularx}{\textwidth}{@{}lYlcccc@{}}
\toprule
Model & Runtime policy & Subset & $n$ &
\begin{tabular}[c]{@{}c@{}}Final label /\\state conn.\end{tabular} &
\begin{tabular}[c]{@{}c@{}}Stable/\\trans.\end{tabular} &
\begin{tabular}[c]{@{}c@{}}Final\\match\end{tabular} \\
\midrule
14B & Threshold similarity & full grid & 189 & $0/189$ / $0/189$ & $0/189$ / $4/189$ & 0.249 \\
14B & State-component bridge & $H=0$ & 9 & $0/9$ / $0/9$ & $0/9$ / $0/9$ & 0.225 \\
14B & State-component bridge & retained & 18 & $7/18$ / $7/18$ & $0/18$ / $18/18$ & 0.798 \\
14B & Label disagreement & $H=0$ & 9 & $0/9$ / $0/9$ & $0/9$ / $1/9$ & 0.259 \\
14B & Label disagreement & retained & 18 & $10/18$ / $10/18$ & $0/18$ / $18/18$ & 0.837 \\
14B & State-distance bridge & $H=0$ & 9 & $0/9$ / $0/9$ & $0/9$ / $3/9$ & 0.254 \\
14B & State-distance bridge & retained & 18 & $8/18$ / $8/18$ & $0/18$ / $18/18$ & 0.828 \\
\addlinespace[2pt]
32B & Threshold similarity & full grid & 189 & $0/189$ / $0/189$ & $0/189$ / $0/189$ & 0.181 \\
32B & State-component bridge & $H=0$ & 9 & $0/9$ / $0/9$ & $0/9$ / $0/9$ & 0.170 \\
32B & State-component bridge & retained & 18 & $18/18$ / $18/18$ & $18/18$ / $18/18$ & 1.000 \\
32B & Label disagreement & $H=0$ & 9 & $0/9$ / $0/9$ & $0/9$ / $0/9$ & 0.180 \\
32B & Label disagreement & retained & 18 & $18/18$ / $18/18$ & $18/18$ / $18/18$ & 1.000 \\
32B & State-distance bridge & $H=0$ & 9 & $0/9$ / $0/9$ & $0/9$ / $0/9$ & 0.207 \\
32B & State-distance bridge & retained & 18 & $18/18$ / $18/18$ & $18/18$ / $18/18$ & 1.000 \\
\bottomrule
\end{tabularx}
\end{table}

Table~\ref{tab:model_scale_controller_app} strengthens both sides of the intervention. Threshold-similarity produces no final behavioral consensus or score-state connectedness in $189$ setting-seed runs for either larger model. At 32B, the retained-memory contrast is exact: every bridge policy reaches final behavioral consensus and score-state connectedness and stable full behavioral consensus in all $18$ runs, while every corresponding $H=0$ run remains terminally fragmented. Qwen2.5-14B shows weaker terminal persistence but the same repair channel: every retained-memory bridge run reaches full behavioral consensus transiently, whereas final consensus is less frequent and no run remains fully aligned throughout the terminal window.

The Qwen2.5 scale contrast does not sharply distinguish the bridge objectives: at Qwen2.5-32B, state-component, label-disagreement, and state-distance routing all repair completely once social evidence is retained. The broader homogeneous comparison below provides the model-family boundary. Retained bridge routing generally repairs both Yi and TinyLlama, but the cleanliness and persistence of that repair remain model conditioned, and threshold-similarity is not universally prohibitive when retained evidence contracts the score geometry sufficiently.

\begin{table}[h]
\caption{Homogeneous TinyLlama-1.1B and Yi-1.5-6B runtime-controller results. Threshold rows aggregate all $H$, $\tau$, seven $\Delta$ values, and seeds $11$--$13$. Bridge rows use $\Delta=0.25$; ``retained'' pools $H\in\{3,10\}$. ``Final cons.'' reports behavioral/score-state connectedness, and ``Stable/trans.'' reports stable and transient full behavioral consensus.}
\label{tab:yi_tiny_controller_app}
\centering
\scriptsize
\setlength{\tabcolsep}{4pt}
\renewcommand{\arraystretch}{1.08}
\begin{tabularx}{\textwidth}{@{}lYlcccc@{}}
\toprule
Model &
Runtime policy &
Subset &
$n$ &
\shortstack{Final label /\\state conn.} &
\shortstack{Stable/\\trans.} &
\shortstack{Final\\match} \\
\midrule
TinyLlama
& Threshold similarity
& full grid
& 189
& $45/189$ / $35/189$
& $38/189$ / $51/189$
& 0.532 \\

TinyLlama
& State-component bridge
& $H=0$
& 9
& $0/9$ / $0/9$
& $0/9$ / $0/9$
& 0.198 \\

TinyLlama
& State-component bridge
& retained
& 18
& $17/18$ / $12/18$
& $10/18$ / $18/18$
& 0.980 \\

TinyLlama
& Label disagreement
& $H=0$
& 9
& $0/9$ / $0/9$
& $0/9$ / $0/9$
& 0.178 \\

TinyLlama
& Label disagreement
& retained
& 18
& $17/18$ / $12/18$
& $13/18$ / $18/18$
& 0.989 \\

TinyLlama
& State-distance bridge
& $H=0$
& 9
& $0/9$ / $0/9$
& $0/9$ / $3/9$
& 0.200 \\

TinyLlama
& State-distance bridge
& retained
& 18
& $16/18$ / $12/18$
& $9/18$ / $18/18$
& 0.968 \\
\addlinespace[2pt]

Yi
& Threshold similarity
& full grid
& 189
& $1/189$ / $1/189$
& $1/189$ / $8/189$
& 0.329 \\

Yi
& State-component bridge
& $H=0$
& 9
& $0/9$ / $0/9$
& $0/9$ / $3/9$
& 0.200 \\

Yi
& State-component bridge
& retained
& 18
& $18/18$ / $18/18$
& $15/18$ / $18/18$
& 1.000 \\

Yi
& Label disagreement
& $H=0$
& 9
& $0/9$ / $0/9$
& $0/9$ / $1/9$
& 0.212 \\

Yi
& Label disagreement
& retained
& 18
& $17/18$ / $18/18$
& $14/18$ / $18/18$
& 0.989 \\

Yi
& State-distance bridge
& $H=0$
& 9
& $0/9$ / $0/9$
& $0/9$ / $0/9$
& 0.247 \\

Yi
& State-distance bridge
& retained
& 18
& $13/18$ / $13/18$
& $10/18$ / $13/18$
& 0.862 \\
\bottomrule
\end{tabularx}
\end{table}

Table~\ref{tab:yi_tiny_controller_app} separates homogeneous coordination ability from heterogeneous compatibility. With retained state-component bridge routing, Yi reaches final behavioral consensus and score-state connectedness in 18/18 settings and TinyLlama reaches them in 17/18 and 12/18. Label-disagreement gives a similar behavioral result, while Yi again reaches score-state connectedness in 18/18 and TinyLlama in 12/18. Yi is therefore not intrinsically unable to coordinate under the paper's protocol; it is at least as repairable homogeneously as TinyLlama and is often cleaner in score state.

Threshold-similarity supplies a complementary boundary condition. Yi remains almost completely fragmented, with only 1/189 final behavioral-consensus and 1/189 final score-state-connected outcomes. TinyLlama reaches final behavioral consensus in 45/189 settings and score-state connectedness in 35/189, all under retained history. The threshold controller is therefore not universally fragmenting by definition. It fragments when model-conditioned score geometry separates before retained evidence can contract the basins sufficiently to preserve cross-basin edges.

\subsection{Consensus attainment versus persistence at 14B and 32B}
\label{app:attainment_persistence}

The attainment-and-persistence comparison uses all 594 homogeneous Qwen2.5-14B and Qwen2.5-32B trajectories to form 297 matched setting pairs. The principal comparison uses the 72 retained-memory non-threshold settings per model, covering well-mixed and all three bridge policies at $H\in\{3,10\}$. Table~\ref{tab:qwen_persistence_app} separates first attainment, subsequent consensus exits, terminal consensus, and final-window stability.

\begin{table}[h]
\caption{Matched retained-memory non-threshold comparison of consensus attainment and persistence. First-consensus state quantities are evaluated at the first full behavioral-consensus round; subsequent quantities are evaluated over the remaining trajectory.}
\label{tab:qwen_persistence_app}
\centering
\scriptsize
\setlength{\tabcolsep}{5pt}
\renewcommand{\arraystretch}{1.10}
\begin{tabularx}{\textwidth}{@{}Xcc@{}}
\toprule
Metric & Qwen2.5-14B & Qwen2.5-32B \\
\midrule
Ever reaches behavioral consensus & 72/72 & 72/72 \\
Final behavioral consensus & 31/72 & 72/72 \\
Stable final-20 behavioral consensus & 0/72 & 72/72 \\
Loses consensus after first attainment & 72/72 & 3/72 \\
Median first-consensus round & 33 & 19.5 \\
Non-consensus fraction after first consensus & 0.534 & $2.5\times10^{-4}$ \\
Mean consensus-exit events & 32.7 & 0.04 \\
Mean state entropy at first consensus & 0.036 & 0.052 \\
Mean top-label probability at first consensus & 0.986 & 0.978 \\
State connected at first consensus & 58/72 & 45/72 \\
Mean state energy after first consensus & 0.737 & 0.0007 \\
Mean temporal JS after first consensus & 0.143 & 0.001 \\
Mean label-change fraction after first consensus & 0.167 & 0.0009 \\
\bottomrule
\end{tabularx}
\end{table}

Matched bootstrap resampling gives a 32B final-consensus advantage of 0.569 with a 95\% interval of approximately $[0.458,0.681]$, a stable-consensus advantage of 1.000, and a reduction in consensus loss after first attainment of 0.958 with interval approximately $[0.903,1.000]$. First consensus occurs about 18.8 rounds earlier at 32B, and post-attainment non-consensus occupancy is lower by about 0.533.

The first-consensus state does not support the interpretation that 14B reaches only weak agreement. Its entropy is lower, its top-label probability is slightly higher, and its state graph is connected more often at first behavioral consensus. The difference emerges after attainment: 14B retains substantial disagreement energy, temporal movement, label changes, and repeated exits, whereas 32B becomes nearly static. The endpoint gap is therefore generated by post-attainment escape rather than by an inability to acquire a convention.

\subsection{Controlled social-evidence order and post-consensus disruption}
\label{app:controlled_social_evidence}

The preceding trajectory analysis shows that Qwen2.5-14B and Qwen2.5-32B both acquire strongly aligned conventions but differ sharply after first attainment. We next use two controlled experiments to test whether this divergence is associated with how the models integrate conflicting social records. The first isolates evidence order at the prompt level. The second transfers the same order manipulation to homogeneous populations that have already acquired a convention.

\subsubsection{Order-controlled social-evidence integration}

The prompt-level assay holds the identity and number of social observations fixed while changing only their order. Histories are written below from oldest to newest. For the three-record 2:1 condition, the minority convention appears first, middle, or last:
\[
(B,A,A),\qquad
(A,B,A),\qquad
(A,A,B).
\]
The five-record 4:1 condition applies the corresponding manipulation with four observations supporting the majority convention and one supporting the minority convention. The 3:2 histories place the two minority records at the oldest positions, at mixed positions, or at the two newest positions.

The assay uses three independently selected tokenizer-safe 10-label vocabularies and five disjoint label pairs within each vocabulary. Each pair is mirrored so that either label serves as the majority convention, preventing a label-specific preference from being mistaken for an order effect. Thirty randomized prompt replicates are evaluated for each directed pair and history template, including randomized allowed-label orders. This gives 900 restricted-score probes per model and history template, 8100 probes per model, and 16,200 probes in total.

For each prompt, let
\[
\ell
=
\log
\frac{p(\text{minority convention})}
{p(\text{majority convention})},
\]
where probabilities are obtained from the full restricted first-token score distribution. The principal contrast is the change in $\ell$ when the identical minority observation is moved from the oldest to the newest position.

\begin{table}[h]
\caption{Order-controlled social-evidence integration. Entries under
`minority top'' are the percentages of 900 probes in which the minority
convention is top-ranked. Histories are ordered from oldest to newest.
$\Delta\ell$ is the mean change in minority-over-majority log odds when the
minority record is moved from oldest to newest. ''Kernel ratio'' is the fitted
newest-position coefficient divided by the oldest-position coefficient for
histories of length three and five.}
\label{tab:controlled_order_app}
\centering
\tiny
\setlength{\tabcolsep}{3.5pt}
\renewcommand{\arraystretch}{1.12}
\begin{tabular}{@{}lccccccccc@{}}
\toprule
&
\multicolumn{4}{c}{2:1 evidence} &
\multicolumn{4}{c}{4:1 evidence} &
\\
\cmidrule(lr){2-5}
\cmidrule(lr){6-9}
Model &
\shortstack{Minority\\oldest} &
\shortstack{Minority\\middle} &
\shortstack{Minority\\newest} &
$\Delta\ell$ &
\shortstack{Minority\\oldest} &
\shortstack{Minority\\middle} &
\shortstack{Minority\\newest} &
$\Delta\ell$ &
\shortstack{Kernel ratio $H=3/H=5$}
\\
\midrule
Qwen2.5-14B
& 0.1\% & 0.3\% & 91.6\% & 23.10
& 0.0\% & 0.6\% & 68.6\% & 21.60
& 3.43 / 4.00
\\

Qwen2.5-32B
& 0.0\% & 0.0\% & 25.0\% & 8.98
& 0.0\% & 0.0\% & 2.7\% & 5.39
& 1.52 / 1.46
\\
\bottomrule
\end{tabular}
\end{table}

Table~\ref{tab:controlled_order_app} shows a large terminal-position effect at both scales but a much weaker accumulated-majority defense at 14B. Under 2:1 evidence, moving the minority observation from oldest to newest increases its mean restricted probability by 0.903 at 14B and 0.276 at 32B. Under 4:1 evidence, the corresponding increases are 0.677 and 0.038. The 14B-minus-32B difference in the newest-minority top-rank rate is positive in all 15 vocabulary--label-pair clusters for both the 2:1 and 4:1 comparisons.

The longer 3:2 histories provide an important qualification. When both minority records occupy the two newest positions, the minority convention becomes top-ranked in 96.3\% of 14B probes and 82.1\% of 32B probes. Thus, 32B is not insensitive to recent evidence. The sharper model distinction is that 32B usually preserves a sufficiently strong numerical majority against a single terminal conflict, whereas 14B frequently does not.

Linear position-weight fits give newest-to-oldest coefficient ratios of 3.43 and 4.00 for 14B at history lengths three and five, compared with 1.52 and 1.46 for 32B. The intermediate coefficients are not perfectly monotonic, so these ratios should be interpreted as summaries of a terminal-position spike rather than evidence for an exact exponential recency kernel. The supported response-level conclusion is that 14B gives disproportionate influence to a final conflicting record and is less protected by accumulated majority evidence.

\subsubsection{Repeated-conflict population intervention}

A population intervention tests whether record position causally changes the stability of an acquired convention. Homogeneous populations contain $N=10$ Qwen2.5-14B or Qwen2.5-32B agents. Each population first receives ten rounds of consistent synthetic social evidence supporting the target convention \texttt{sol}. During the following ten intervention rounds, a fixed subset of agents receives one of three three-record histories:
\[
\begin{array}{ll}
\text{No conflict:}     & (\texttt{sol},\texttt{sol},\texttt{sol}), \\
\text{Oldest conflict:} & (\texttt{vel},\texttt{sol},\texttt{sol}), \\
\text{Newest conflict:} & (\texttt{sol},\texttt{sol},\texttt{vel}).
\end{array}
\]

The oldest- and newest-conflict conditions contain identical labels and
evidence counts. Only the location of the conflicting observation differs.

Conflict is applied to 10\%, 30\%, or 50\% of the agents, with seeds
11--13, under well-mixed and state-component bridge routing. A shared
no-conflict run is included for every model, policy, and seed. The design
therefore contains
\[
\begin{aligned}
&2\text{ models}
\times 2\text{ policies} \\
&\quad\times
\left(
2\text{ conflict positions}
\times 3\text{ affected fractions}
\times 3\text{ seeds}
+
3\text{ control seeds}
\right)
=84
\end{aligned}
\]
population runs. Artificial conflict is removed after the intervention, followed by 80 clean recovery rounds.

A run ``loses the target convention'' when at least one post-warmup round no longer has all ten agents emitting \texttt{sol}. The peak conflicting count is the maximum number of agents emitting \texttt{vel} during intervention. State energy is averaged over the ten intervention rounds.

\begin{table}[h]
\caption{Routing-resolved repeated-conflict intervention. Newest and oldest conditions each pool three affected fractions and three seeds, giving nine runs per cell; controls contain three seeds. ''Loss'' counts runs that leave full target consensus after warmup. ''Peak conflict'' and ``State energy'' report newest/oldest means during intervention. Every row contains 21 total runs, all of which finish in target consensus and satisfy stable target consensus throughout the final 20 rounds.}
\label{tab:repeated_conflict_app}
\centering
\scriptsize
\setlength{\tabcolsep}{5pt}
\renewcommand{\arraystretch}{1.12}
\begin{tabular}{@{}llccccc@{}}
\toprule
Model & Policy &
\shortstack{Newest\\loss} &
\shortstack{Oldest\\loss} &
\shortstack{Control\\loss} &
\shortstack{Peak conflict\\newest/oldest} &
\shortstack{State energy\\newest/oldest} \\
\midrule
14B & Well mixed
& 9/9 & 9/9 & 0/3
& 2.22 / 1.44
& 0.581 / 0.234 \\

14B & State-component
& 9/9 & 1/9 & 0/3
& 2.22 / 0.11
& 0.694 / 0.002 \\

32B & Well mixed
& 9/9 & 0/9 & 0/3
& 1.67 / 0.00
& 0.323 / 0.000 \\

32B & State-component
& 9/9 & 0/9 & 0/3
& 2.11 / 0.00
& 0.535 / 0.000 \\
\bottomrule
\end{tabular}
\end{table}

All 84 populations establish the target convention by the end of warmup. Pooling routing policies and affected fractions, newest conflict causes a target-consensus loss in 18/18 runs at 14B and 18/18 at 32B. Oldest conflict causes a loss in 10/18 runs at 14B and 0/18 at 32B, while no-conflict controls remain aligned in 12/12 runs.

The routing-resolved contrast further qualifies the 14B response. Under state-component bridge routing, moving the conflict from newest to oldest reduces target-consensus loss from 9/9 to 1/9 and reduces mean intervention state energy from 0.694 to 0.002. Under well-mixed routing, oldest conflict still causes a behavioral loss in 9/9 14B runs, although its mean peak conflicting count and state energy remain smaller than under newest conflict. At 32B, oldest conflict produces no sampled-label adoption under either policy.

Disruption also increases with the affected fraction. Pooling the two routing policies, the mean peak number of conflicting agents under newest conflict rises from 1.00 at 10\% exposure to 2.50 at 30\% and 3.17 at 50\% for 14B. The corresponding 32B means are 1.00, 1.67, and 3.00. This dose response shows that the controlled record-position effect scales to a collective disturbance rather than remaining only a single-agent score shift.

The recovery phase supplies an equally important negative result. Every run returns to the target convention, finishes in behavioral consensus with a connected score-state graph, and remains in target consensus throughout the final 20 rounds. Repeated newest conflict can therefore disrupt either model when imposed strongly enough, including Qwen2.5-32B. The intervention establishes that record position causally changes transient population disruption, but it does not reproduce the persistent endogenous maintenance difference in Table~\ref{tab:qwen_persistence_app}.

Together, the two experiments support a bounded mechanism claim. Qwen2.5-14B displays a substantially stronger terminal-position effect and weaker resistance to a final minority observation than Qwen2.5-32B. The same order manipulation has causal collective consequences after consensus has been acquired. These findings make order-sensitive social-evidence integration a plausible contributor to the repeated endogenous exits at 14B, but they do not show that this response property alone is sufficient to generate its full long-run persistence regime.

\subsection{State-similarity threshold robustness}
\label{app:delta_robustness}

The main text uses $\Delta=0.25$ as the reporting and state-similarity threshold. Because state components are defined by thresholding Jensen--Shannon distances, we test whether the qualitative conclusions depend on this choice by evaluating
\[
    \Delta\in\{0.15,0.20,0.25,0.30,0.35\}
\]
over seeds $11$--$13$.

The role of $\Delta$ differs across controllers. For threshold-similarity routing and state-component bridge routing, $\Delta$ enters runtime edge construction and can therefore change the behavioral trajectory. For well-mixed, label-disagreement, and state-distance routing, $\Delta$ does not change the behavioral routing rule; it only changes the diagnostic state-similarity graph and quantities derived from that graph. Table~\ref{tab:delta_robustness_app} therefore distinguishes unique stochastic trajectories from setting--seed--threshold diagnostic evaluations. We do not attach independence-based confidence intervals to repeated diagnostic evaluations of the same trajectory.

\begin{table}[h]
\caption{Robustness to the state-similarity threshold $\Delta$. ``Traj.'' is the number of unique stochastic trajectories and ``Eval.'' is the number of setting--seed--threshold evaluations. For well-mixed, label-disagreement, and state-distance routing, changing $\Delta$ re-evaluates the same behavioral trajectory under a different diagnostic state threshold; for threshold-similarity and state-component bridge routing, it can also change runtime pairings. Slash-separated entries report final behavioral consensus / score-state connectedness and mean label/state component counts.}
\label{tab:delta_robustness_app}
\centering
\scriptsize
\setlength{\tabcolsep}{4pt}
\renewcommand{\arraystretch}{1.12}
\begin{tabularx}{\textwidth}{@{}Xccccc@{}}
\toprule
Setting family &
Traj. &
Eval. &
\begin{tabular}[c]{@{}c@{}}Final behavioral /\\state conn.\end{tabular} &
Final match &
\begin{tabular}[c]{@{}c@{}}Mean comps.\\label/state\end{tabular} \\
\midrule
Qwen2.5-7B well-mixed, $H=0,\tau=0.3$
& 3 & 15 & $0/15$ / $0/15$ & 0.259 & 3.67 / 3.73 \\

Qwen2.5-7B well-mixed, $H=3,\tau=0.3$
& 3 & 15 & $15/15$ / $15/15$ & 1.000 & 1.00 / 1.00 \\

Mixed threshold-similarity, $H=3,\tau=0.3$
& 15 & 15 & $0/15$ / $0/15$ & 0.173 & 4.67 / 7.73 \\

Mixed state-component bridge, $H=3,\tau=0.3$
& 15 & 15 & $12/15$ / $8/15$ & 0.939 & 1.20 / 1.87 \\

Mixed label disagreement, $H=3,\tau=0.3$
& 3 & 15 & $15/15$ / $7/15$ & 1.000 & 1.00 / 1.67 \\

Mixed state-distance bridge, $H=3,\tau=0.3$
& 3 & 15 & $0/15$ / $0/15$ & 0.467 & 2.00 / 2.00 \\

Mixed bridge variants, $H=0,\tau=0.1$
& 21 & 45 & $0/45$ / $0/45$ & 0.152 & 5.18 / 7.47 \\
\bottomrule
\end{tabularx}
\end{table}

Table~\ref{tab:delta_robustness_app} supports the same qualitative mechanism as the main text. Changing $\Delta$ does not turn no-memory settings into consensus, and threshold-similarity remains fragmented across the runtime thresholds tested. State-component bridge gives strong but threshold-sensitive behavioral and state-space repair. Label-disagreement produces behavioral consensus in each of its three unique seed trajectories, while its reported state structure changes as the diagnostic threshold changes. State-distance bridge remains weak. The main conclusion is therefore not based on treating repeated diagnostic thresholds as independent runs: homophilous threshold-similarity remains unreliable, while retained-memory bridge policies that expose relevant basins produce substantially stronger repair.

\subsection{Population-size stress tests}
\label{app:population_size_robustness}

The population-size contrast evaluates whether the graph and memory mechanisms persist at $N\in\{20,50\}$. Table~\ref{tab:population_size_stress_app} pools seeds 11--13 at both sizes. All runs use $T=200$ rounds and $\Delta=0.25$. Homogeneous Qwen2.5-7B well-mixed populations use $\tau=0.3$; heterogeneous retained-memory controller runs use $H=3,\tau=0.3$; and the no-memory row pools the three bridge-seeking controllers at $H=0,\tau=0.1$. The $N=10$ experiments provide inspectable graph artifacts, while the larger populations test whether the same mechanisms persist under a fixed interaction budget.

\begin{table}[h]
\caption{Population-size stress tests at $N=20$ and $N=50$. Counts aggregate over seeds $11$--$13$ and both population sizes unless otherwise stated. ``Stable/trans.'' gives stable and transient full behavioral consensus counts. Match intervals are bootstrap 95\% intervals over setting rows. Slash-separated entries report final behavioral consensus / score-state connectedness and mean label/state component counts.}
\label{tab:population_size_stress_app}
\centering
\scriptsize
\setlength{\tabcolsep}{4pt}
\renewcommand{\arraystretch}{1.12}
\begin{tabularx}{\textwidth}{@{}Xccccc c@{}}
\toprule
Setting & $n$ &
\begin{tabular}[c]{@{}c@{}}Final label /\\state conn.\end{tabular} &
Stable/trans. & Final match &
\begin{tabular}[c]{@{}c@{}}Mean comps.\\label/state\end{tabular} &
Full frac. \\
\midrule
Qwen2.5-7B well-mixed, $H=0$ & 6 & $0/6$ / $0/6$ & $0/6$ / $0/6$ & $0.144$ $[0.110,0.179]$ & 8.00 / 9.17 & $0.000$ \\
Qwen2.5-7B well-mixed, $H=3$ & 6 & $6/6$ / $6/6$ & $6/6$ / $6/6$ & $1.000$ $[1.000,1.000]$ & 1.00 / 1.00 & $0.942$ \\
Qwen2.5-7B well-mixed, $H=10$ & 6 & $6/6$ / $6/6$ & $6/6$ / $6/6$ & $1.000$ $[1.000,1.000]$ & 1.00 / 1.00 & $0.922$ \\
Mixed threshold-similarity, $H=3$ & 6 & $0/6$ / $0/6$ & $0/6$ / $0/6$ & $0.159$ $[0.131,0.186]$ & 8.33 / 19.83 & $0.000$ \\
Mixed state-component bridge, $H=3$ & 6 & $2/6$ / $3/6$ & $0/6$ / $6/6$ & $0.856$ $[0.760,0.952]$ & 1.67 / 2.00 & $0.367$ \\
Mixed label-disagreement, $H=3$ & 6 & $1/6$ / $1/6$ & $1/6$ / $6/6$ & $0.769$ $[0.658,0.891]$ & 1.83 / 2.83 & $0.297$ \\
Mixed state-distance bridge, $H=3$ & 6 & $1/6$ / $1/6$ & $0/6$ / $1/6$ & $0.578$ $[0.483,0.749]$ & 1.83 / 2.33 & $0.138$ \\
Mixed bridge-seeking controllers, $H=0$ & 18 & $0/18$ / $0/18$ & $0/18$ / $0/18$ & $0.145$ $[0.129,0.162]$ & 8.22 / 16.28 & $0.000$ \\
\bottomrule
\end{tabularx}
\end{table}

The population-size contrast preserves the qualitative mechanism and identifies where the current controller becomes finite-budget limited. Retained history remains decisive in homogeneous Qwen2.5-7B populations: at both $N=20$ and $N=50$, $H=0$ fragments, whereas $H=3$ and $H=10$ reach final behavioral consensus and score-state connectedness in every seed. Mixed threshold-similarity remains strongly fragmented, with no final or transient full consensus and a nearly always-disconnected endogenous runtime graph.

The bridge-seeking results are more nuanced. At larger population sizes, state-component bridge and label-disagreement routing still substantially improve match rates and induce transient full consensus in every retained-memory run, but terminal stable repair is no longer reliable at the fixed $T=200$ budget. State-component bridge gives the strongest retained-memory match rate and the cleanest average state-space repair among the bridge controllers, while label-disagreement gives strong behavioral pressure but more residual state structure. State-distance bridge remains weaker and produces little transient consensus. The population-size tests therefore characterize a fixed-budget controller: the qualitative mechanism persists, while reliable terminal repair at larger $N$ may require more rounds, larger bridge budgets, adaptive stopping, or population-size-aware routing.

\subsection{Vocabulary-size robustness}
\label{app:vocab_robustness}

The main text uses a fixed allowed-label vocabulary size. We test whether the qualitative conclusions depend on this choice by sweeping tokenizer-safe nonce vocabularies with
\[
    M\in\{5,10,20\}.
\]
All runs use $N=10$, $T=200$, $\Delta=0.25$, and seeds $11$--$13$. Homogeneous Qwen2.5-7B well-mixed runs compare $H=0$ and $H=3$ at $\tau=0.3$. Mixed four-model graph-control runs use $H=3,\tau=0.3$ and compare threshold-similarity, state-component bridge, label-disagreement, and state-distance bridge routing. Table~\ref{tab:vocab_robustness_app} reports pooled results over the three vocabulary sizes and three seeds.

\begin{table}[h]
\caption{Vocabulary-size robustness over tokenizer-safe nonce vocabularies with $M\in\{5,10,20\}$. Counts aggregate over three vocabulary sizes and seeds $11$--$13$. ``Stable/trans.'' gives stable and transient full behavioral consensus counts. Match intervals are bootstrap 95\% intervals over setting rows. Slash-separated entries report final behavioral consensus / score-state connectedness and mean label/state component counts.}
\label{tab:vocab_robustness_app}
\centering
\scriptsize
\setlength{\tabcolsep}{4pt}
\renewcommand{\arraystretch}{1.12}
\begin{tabularx}{\textwidth}{@{}Xccccc c@{}}
\toprule
Setting & $n$ &
\begin{tabular}[c]{@{}c@{}}Final label /\\state conn.\end{tabular} &
Stable/trans. & Final match &
\begin{tabular}[c]{@{}c@{}}Mean comps.\\label/state\end{tabular} &
Full frac. \\
\midrule
Qwen2.5-7B well-mixed, $H=0$ & 9 & $0/9$ / $0/9$ & $0/9$ / $2/9$ & $0.254$ $[0.185,0.333]$ & 3.89 / 3.89 & $0.001$ \\
Qwen2.5-7B well-mixed, $H=3$ & 9 & $9/9$ / $9/9$ & $9/9$ / $9/9$ & $1.000$ $[1.000,1.000]$ & 1.00 / 1.00 & $0.943$ \\
Mixed threshold-similarity, $H=3$ & 9 & $0/9$ / $0/9$ & $0/9$ / $0/9$ & $0.163$ $[0.123,0.200]$ & 4.89 / 7.89 & $0.000$ \\
Mixed state-component bridge, $H=3$ & 9 & $8/9$ / $5/9$ & $4/9$ / $9/9$ & $0.978$ $[0.933,1.000]$ & 1.11 / 1.78 & $0.744$ \\
Mixed label-disagreement, $H=3$ & 9 & $9/9$ / $5/9$ & $3/9$ / $9/9$ & $1.000$ $[1.000,1.000]$ & 1.00 / 1.44 & $0.807$ \\
Mixed state-distance bridge, $H=3$ & 9 & $2/9$ / $1/9$ & $1/9$ / $3/9$ & $0.622$ $[0.501,0.780]$ & 1.89 / 2.22 & $0.182$ \\
\bottomrule
\end{tabularx}
\end{table}

Table~\ref{tab:vocab_robustness_app} supports the same mechanism as the main text. The homogeneous memory transition is invariant across the tested vocabulary sizes: $H=0$ remains fragmented, while $H=3$ reaches final behavioral consensus and score-state connectedness for every seed and vocabulary size. Mixed threshold-similarity routing remains strongly fragmented, with no transient or stable full-consensus run and an almost always-disconnected endogenous runtime graph.

The bridge-seeking results also preserve the qualitative ordering. State-component bridge and label-disagreement routing both produce strong retained-memory repair across vocabularies, while state-distance bridge remains much weaker. Label-disagreement is behaviorally strongest in the vocabulary-size contrast, reaching final label consensus in all pooled runs, but both component bridge and label-disagreement can leave residual state structure or fail the terminal stability criterion in some seeds. Thus, vocabulary size does not reverse the main conclusion: useful repair depends on retained memory and on a bridge objective that exposes the relevant convention basins, whereas homophilous threshold-similarity and generic distance pairing remain unreliable.

\section{Population-composition contrasts}
\label{app:composition_boundary}

The population-composition study evaluates five $N=10$ cohorts under a shared protocol. The cohorts contain: (i) three Qwen2.5-32B, three Qwen2.5-14B, two Yi-1.5-6B, and two TinyLlama agents; (ii) three Qwen2.5-32B, three Qwen2.5-14B, and four Qwen2.5-7B agents; (iii) the same 32B/14B/7B Qwen backbone with two 7B positions occupied by TinyLlama; (iv) the same backbone with those positions occupied by Yi; and (v) three Qwen2.5-32B, three Qwen2.5-7B, two Yi, and two TinyLlama agents. Together, these compositions isolate scale heterogeneity, TinyLlama-versus-Yi family substitution, and the presence of 14B agents. All runs use $M=10$, $T=200$, $\tau=0.3$, $\Delta=0.25$, seeds 11--13, and $H\in\{0,3\}$. Each composition is evaluated under well-mixed, threshold-similarity, state-component, label-disagreement, and state-distance routing, giving 150 successful runs.

Across all 75 no-memory runs, no policy produces final behavioral consensus, final score-state connectedness, stable final-window behavioral consensus, or even transient full consensus. Retained-memory threshold-similarity likewise produces none of these outcomes in any of 15 runs. Table~\ref{tab:large_boundary_app} reports the retained-memory non-threshold outcomes.

\begin{table}[h]
\caption{Retained-memory large heterogeneous cohorts at $H=3$, $\tau=0.3$, $\Delta=0.25$, and three seeds. Each entry gives final behavioral consensus / final score-state connectedness / stable final-20 behavioral consensus / ever-reached full behavioral consensus. Threshold-similarity is 0/3 for all four outcomes in every cohort.}
\label{tab:large_boundary_app}
\centering
\scriptsize
\setlength{\tabcolsep}{5pt}
\renewcommand{\arraystretch}{1.10}
\begin{tabularx}{\textwidth}{@{}p{0.30\textwidth}cccc@{}}
\toprule
Cohort & State-component & Label-disagreement & State-distance & Well mixed \\
\midrule
Qwen 32B/14B + Yi + TinyLlama
& 0/0/0/0 & 0/0/0/0 & 0/0/0/0 & 0/0/0/0 \\
Qwen2.5 32B/14B/7B
& 3/3/3/3 & 3/3/3/3 & 3/3/3/3 & 3/3/3/3 \\
Qwen 32B/14B/7B + TinyLlama
& 3/3/3/3 & 3/3/3/3 & 0/0/0/0 & 3/3/3/3 \\
Qwen 32B/14B/7B + Yi
& 0/0/0/2 & 0/0/0/2 & 0/0/0/2 & 1/0/0/3 \\
Qwen 32B/7B + Yi + TinyLlama
& 0/0/0/0 & 0/0/0/0 & 0/0/0/0 & 0/0/0/0 \\
\bottomrule
\end{tabularx}
\end{table}

The same-family Qwen cohort shows that scale heterogeneity alone does not prevent repair: every non-threshold policy reaches final behavioral consensus and score-state connectedness and remains stable in all three seeds. The Qwen--TinyLlama cohort likewise repairs under state-component, label-disagreement, and well-mixed routing, while state-distance fails in all seeds. Replacing the two TinyLlama agents with Yi changes the result despite preserving the large Qwen backbone. State-component, label-disagreement, and state-distance routing each attain transient consensus in two of three seeds but never finish in behavioral consensus or with a connected score-state graph; well mixed attains transient consensus in all seeds, finishes behaviorally in one, and never finishes with a connected score-state graph.

The Qwen 32B/14B + Yi + TinyLlama population never attains full consensus under any retained-memory policy. Removing 14B does not restore repair. In that no-14B cohort, state-component, label-disagreement, and well-mixed routing terminate at \texttt{sol:8, vel:2} in every seed: Qwen and TinyLlama agents use \texttt{sol}, while the two Yi agents use \texttt{vel}. State-distance is more fragmented, ending in two or three sampled-label components rather than reproducing the clean family-aligned split. This qualification is important: the family-aligned endpoint is repeated under the policies that provide broad cross-basin exposure, not under every non-threshold policy.

The homogeneous controls in Supplementary Section~B.2 rule out the interpretation that the Yi-containing mixed failures arise because Yi cannot coordinate under the protocol. Under retained state-component bridge routing, homogeneous Yi reaches final behavioral consensus and score-state connectedness in 18/18 settings and stable final-window consensus in 15/18. Homogeneous TinyLlama reaches final behavioral consensus in 17/18, score-state connectedness in 12/18, and stable consensus in 10/18, yet the matched Qwen--TinyLlama cohort repairs while the Qwen--Yi cohort does not. Homogeneous repairability therefore does not determine cross-family compatibility.

\subsection{Geometry, exposure, and family-resolved adoption}
\label{app:family_adoption}

Terminal Yi and non-Yi score states in the no-14B cohort are separated by observed Jensen--Shannon distances of approximately 0.70--0.83. These distances exceed both the threshold $\Delta=0.25$ and the state-distance controller's upper limit of 0.50. Thus, the corresponding terminal cross-family pairs are ineligible under both routing rules. It does not explain failure under state-component, label-disagreement, or well-mixed routing, which repeatedly expose Yi and non-Yi agents.

For a counted directed event, a Yi and non-Yi agent disagree when the source contribution is observed, and adoption means that the target emits the source's previous label at the next round. Table~\ref{tab:family_adoption_app} shows that exposure is abundant but cross-family adoption is rare.

\begin{table}[h]
\caption{Directed cross-family adoption in the no-14B cohort. Percentages are adoption after a qualifying Yi/non-Yi disagreement exposure; parentheses give qualifying event counts across three seeds.}
\label{tab:family_adoption_app}
\centering
\small
\setlength{\tabcolsep}{6pt}
\begin{tabular}{lcc}
\toprule
Policy & Yi adopts non-Yi & Non-Yi adopts Yi \\
\midrule
State-component bridge & 1.5\% (1157) & 0.4\% (1125) \\
Label disagreement & 3.0\% (1068) & 5.7\% (1046) \\
Well mixed & 2.0\% (1083) & 0.6\% (1017) \\
\bottomrule
\end{tabular}
\end{table}

The low rates are not a small-sample artifact: each reported direction contains roughly one thousand qualifying events per policy. The result separates graph accessibility from response compatibility. A controller can repeatedly create cross-family exposure without producing durable basin merging.

\subsection{Stationary priors and synthetic-history susceptibility}
\label{app:susceptibility_probes}

The susceptibility contrast contains 450 stationary-prior probes, 900 non-empty synthetic-history probes, and 12 seeded population runs, for 1362 probes or runs. Stationary probes use three independently selected tokenizer-safe vocabularies and 30 randomized label orders per model--vocabulary pair. The history campaign uses six non-empty three-record conditions per model: consistent \texttt{sol}, consistent \texttt{vel}, and four mixed 2:1 templates.

\begin{table}[h]
\caption{Stationary restricted-score concentration and response to three consistent rendered history records. Prior entropy is averaged over the three vocabularies. History columns count randomized-order probes in which the displayed convention becomes top-ranked.}
\label{tab:susceptibility_consistent_app}
\centering
\small
\setlength{\tabcolsep}{6pt}
\begin{tabular}{lccc}
\toprule
Model & Prior entropy & \texttt{sol}$\times3$ & \texttt{vel}$\times3$ \\
\midrule
Qwen2.5-32B & 0.074 & 30/30 & 30/30 \\
Qwen2.5-14B & 0.218 & 30/30 & 30/30 \\
Qwen2.5-7B & 0.052 & 30/30 & 30/30 \\
Yi-1.5-6B & 1.369 & 15/30 & 22/30 \\
TinyLlama-1.1B & 1.390 & 30/30 & 0/30 \\
\bottomrule
\end{tabular}
\end{table}

Yi's stationary prior is diffuse and changes top label across orderings and vocabularies, so the recurrent terminal \texttt{vel} basin is not explained by a fixed intrinsic \texttt{vel} preference. The consistent-history results instead reveal partial and asymmetric responsiveness. All Qwen variants can be driven to either repeated convention. Yi follows the displayed convention in only half of the \texttt{sol} probes and 22/30 \texttt{vel} probes. TinyLlama follows \texttt{sol} in every probe and never makes repeated \texttt{vel} top-ranked. Identical rendered evidence can therefore create opposing response basins in a Yi--TinyLlama population.

Mixed 2:1 histories provide a broader comparison across the homogeneous Qwen2.5 scales. Across four templates, the repeated convention remains top-ranked in 87.5\% of Qwen2.5-32B probes, 7.5\% of 14B probes, and 0.8\% of 7B probes. For two \texttt{sol} records followed by conflicting \texttt{vel}, the corresponding rates are 90\%, 20\%, and 3.3\%; for the mirrored conflict they are 83.3\%, 0\%, and 0\%. These coarse susceptibility probes show that the response regime is not monotonic in model size and that isolated history following is not by itself a scalar predictor of population persistence. Supplementary Section~\ref{app:controlled_social_evidence} therefore isolates the 14B/32B maintenance contrast using matched evidence-order permutations and a post-consensus population intervention.

The model priors used for routing normalization in Supplementary Sec.~\ref{app:prior_normalized_routing} are separate prompt- and vocabulary-conditioned estimates on the exact mixed-cohort ten-label vocabulary. The three-vocabulary stationary and synthetic-history probes in this section serve the broader diagnostic purpose of comparing model susceptibility and should not be interpreted as the routing priors consumed by the normalized controllers.

The seeded-population control initializes every agent in the Qwen 32B/14B + Yi + TinyLlama cohort with three synthetic \texttt{sol} or three synthetic \texttt{vel} records before normal routing resumes. State-component and well-mixed routing are each evaluated for three seeds under both initial conventions. None of the 12 runs reaches final behavioral consensus, final score-state connectedness, or stable final-20 behavioral consensus. All-\texttt{sol} seeding under well mixed produces transient consensus in all three seeds, but none maintains it to the end. Initial conditions can change the trajectory without eliminating the composition-dependent failure.

\section{Task-grounded answer-choice ablations}
\label{app:task_choice_ablations}

The label-only study uses homogeneous populations of $N=6$ Qwen2.5-7B-Instruct agents for $T=20$ rounds, with retained history $H=3$, state threshold $\Delta=0.25$, temperatures $\tau\in\{0.3,0.7\}$, and seeds 11--13. The four routing policies are well mixed, threshold similarity, state-component bridge, and label disagreement. Agents choose only from answer labels $\{\texttt{A},\texttt{B},\texttt{C},\texttt{D}\}$ and exchange labels through local histories; no rationale text is generated or inserted.

ARC-Challenge uses the complete validation subset with exactly four choices, leaving 295 of 299 items. Each ARC controller cell therefore contains $295\times2\times3=1770$ item--temperature--seed runs. MMLU-100 uses a fixed 100-item manifest, giving $100\times2\times3=600$ runs per controller cell. The manifest construction and selection criterion are reported in Supplementary Sec.~H.

\begin{table}[h]
\caption{Task-grounded label-only graph-controller validation. ARC-Challenge uses the fixed-four-choice filtered validation set ($295$ items); MMLU uses $100$ questions. Rates are averaged over item--seed--temperature runs. ``Final corr.'' is the final per-agent correctness rate. ``Maj. corr.'' indicates whether the final majority label is the gold answer. ``Correct cons.'' and ``Wrong cons.'' split final behavioral consensus by whether the consensus label is correct. ``State conn.'' is final score-state connectedness of the
restricted answer-choice distributions. $W{\to}C$ and $C{\to}W$ are agent-level wrong-to-correct and correct-to-wrong changes from round 0 to the final round.}
\label{tab:task_choice_label_only_app}
\centering
\tiny
\setlength{\tabcolsep}{3pt}
\renewcommand{\arraystretch}{1.12}
\begin{tabular}{@{}llrrrrrrrr@{}}
\toprule
Task & Runtime policy & $n$ &
\begin{tabular}[c]{@{}c@{}}Final\\corr.\end{tabular} &
\begin{tabular}[c]{@{}c@{}}Maj.\\corr.\end{tabular} &
\begin{tabular}[c]{@{}c@{}}Correct\\cons.\end{tabular} &
\begin{tabular}[c]{@{}c@{}}Wrong\\cons.\end{tabular} &
\begin{tabular}[c]{@{}c@{}}State\\conn.\end{tabular} &
$W{\to}C$ & $C{\to}W$ \\
\midrule
ARC-Challenge & State-component bridge & 1770 & 0.881 & 0.888 & 0.834 & 0.071 & 0.901 & 0.024 & 0.027 \\
ARC-Challenge & Label disagreement & 1770 & 0.878 & 0.885 & 0.832 & 0.069 & 0.894 & 0.022 & 0.027 \\
ARC-Challenge & Threshold-similarity & 1770 & 0.870 & 0.880 & 0.803 & 0.055 & 0.848 & 0.019 & 0.027 \\
ARC-Challenge & Well-mixed & 1770 & 0.875 & 0.881 & 0.823 & 0.074 & 0.894 & 0.024 & 0.032 \\
\addlinespace[2pt]
MMLU-100 & State-component bridge & 600 & 0.336 & 0.338 & 0.125 & 0.295 & 0.567 & 0.153 & 0.152 \\
MMLU-100 & Label disagreement & 600 & 0.341 & 0.342 & 0.145 & 0.272 & 0.563 & 0.155 & 0.149 \\
MMLU-100 & Threshold-similarity & 600 & 0.333 & 0.347 & 0.080 & 0.180 & 0.330 & 0.148 & 0.150 \\
MMLU-100 & Well-mixed & 600 & 0.339 & 0.337 & 0.138 & 0.278 & 0.573 & 0.163 & 0.160 \\
\bottomrule
\end{tabular}
\end{table}

Table~\ref{tab:task_choice_label_only_app} shows that the graph-control lens transfers from nonce conventions to task-grounded answer choices. On filtered ARC-Challenge validation, correctness is already high, so the main effect is not a large accuracy shift. Instead, topology changes how often populations settle into behavioral consensus and score-state connectedness. Threshold-similarity has lower correct consensus and score-state connectedness than the bridge-seeking and well-mixed policies. State-component bridge gives the highest final correctness, majority correctness, correct consensus, and score-state connectedness in this suite. MMLU-100 gives a more balanced picture: final correctness is similar across controllers, but threshold-similarity suppresses both behavioral consensus and score-state connectedness relative to bridge and well-mixed policies. Thus, the task-grounded results support the propagation lens while ruling out the stronger claim that bridge routing is a general accuracy-improvement method.

To look beyond final correctness, we also compute strict-majority flips from round 0 to the final round. A run is counted as ``round-0 wrong majority $\to$ final correct majority'' only if round 0 has a strict majority for a non-gold answer and the final round has a strict majority for the gold answer. A run is counted as ``round-0 correct majority $\to$ final wrong majority'' only if the direction reverses. Runs with no strict majority at either endpoint are not counted as flips.

\begin{table}[h]
\caption{Strict-majority flips in the label-only task-choice study. Counts are over all item--seed--temperature runs in each task-controller cell. $W_{\mathrm{maj}}{\to}C_{\mathrm{maj}}$ is round-0 wrong strict majority to final correct strict majority. $C_{\mathrm{maj}}{\to}W_{\mathrm{maj}}$ is round-0 correct strict majority to final wrong strict majority.}
\label{tab:task_choice_majority_flips_label_only_app}
\centering
\tiny
\setlength{\tabcolsep}{4pt}
\renewcommand{\arraystretch}{1.12}
\begin{tabular}{@{}llrrrrr@{}}
\toprule
Task & Runtime policy & $n$ &
\begin{tabular}[c]{@{}c@{}}Round-0\\strict maj.\end{tabular} &
\begin{tabular}[c]{@{}c@{}}Final\\strict maj.\end{tabular} &
$W_{\mathrm{maj}}{\to}C_{\mathrm{maj}}$ &
$C_{\mathrm{maj}}{\to}W_{\mathrm{maj}}$ \\
\midrule
ARC-Challenge & State-component bridge & 1770 & 0.986 & 0.980 & $30/1770$ & $17/1770$ \\
ARC-Challenge & Label disagreement & 1770 & 0.986 & 0.972 & $24/1770$ & $20/1770$ \\
ARC-Challenge & Threshold-similarity & 1770 & 0.980 & 0.949 & $20/1770$ & $11/1770$ \\
ARC-Challenge & Well-mixed & 1770 & 0.989 & 0.972 & $26/1770$ & $29/1770$ \\
\addlinespace[2pt]
MMLU-100 & State-component bridge & 600 & 0.825 & 0.798 & $54/600$ & $46/600$ \\
MMLU-100 & Label disagreement & 600 & 0.825 & 0.815 & $55/600$ & $44/600$ \\
MMLU-100 & Threshold-similarity & 600 & 0.798 & 0.730 & $36/600$ & $38/600$ \\
MMLU-100 & Well-mixed & 600 & 0.833 & 0.802 & $54/600$ & $49/600$ \\
\bottomrule
\end{tabular}
\end{table}

The strict-majority analysis gives the same cautious interpretation. On ARC-Challenge, structured bridge policies have more wrong-majority-to-correct-majority flips than the reverse, but the counts are small relative to the high initial correctness of the filtered validation set. On MMLU-100, flips are more balanced: state-component bridge, label-disagreement, and well-mixed have slightly more wrong-to-correct majority flips, while threshold-similarity is nearly neutral in the opposite direction. These results show that graph topology changes answer propagation, but the direction and usefulness of propagation depend on the task distribution and initial answer basins.

\subsection{Rationale-sharing contrast on the full task manifests}
\label{app:rationale_sharing}

The rationale-sharing contrast uses the same fixed task manifests as the task-grounded study: all $295$ fixed-four-choice ARC-Challenge validation items and the fixed MMLU-100 manifest. All conditions use Qwen2.5-7B-Instruct populations with six agents, seed $11$, $\tau=0.3$, $H=3$, $\Delta=0.25$, and $T=20$ rounds. Each condition is evaluated under well-mixed, threshold-similarity, state-component, and label-disagreement routing. An ARC row therefore aggregates $295\times4=1180$ item--policy runs, while an MMLU row aggregates $100\times4=400$ item--policy runs.

The conditions isolate when rationales are generated and whether their semantic association with the source answer is preserved. In the \emph{posthoc rationale} condition, an agent first selects its answer and then generates a short rationale conditioned on that answer for use in subsequent interaction histories. In the \emph{rationale-first} condition, the rationale is generated before the final answer. The \emph{rationale-first with shuffled rationales} condition preserves generated text and prompt length while breaking the correspondence between a rationale and the answer source from which it was generated. All generated rationales are nonempty, and no rationale-generation failures are recorded in the aggregated runs.

\begin{table}[h]
\caption{Rationale-sharing results on the full ARC-295 and MMLU-100 manifests. Rows pool the four routing policies at seed $11$, $\tau=0.3$, $H=3$, and $T=20$. ``Maj. corr.'' indicates that the final dominant label is correct. ``Beh. cons.'' is final behavioral consensus. The final column reports mean final label/state component counts.}
\label{tab:task_choice_rationale_app}
\centering
\tiny
\setlength{\tabcolsep}{3.2pt}
\renewcommand{\arraystretch}{1.12}
\begin{tabularx}{\textwidth}{@{}lXrcccccc@{}}
\toprule
Task &
Condition &
$n$ &
\shortstack{Final\\corr.} &
\shortstack{Maj.\\corr.} &
\shortstack{Beh.\\cons.} &
\shortstack{Correct\\conn.} &
\shortstack{Wrong\\cons.} &
\shortstack{Mean comps.\\label/state} \\
\midrule
ARC-295
& Posthoc rationale
& 1180 & .877 & .886 & .903 & .829 & .075 & 1.123/1.212 \\

ARC-295
& Rationale first
& 1180 & .880 & .887 & .894 & .826 & .068 & 1.130/1.197 \\

ARC-295
& Rationale first + shuffled
& 1180 & .887 & .897 & .874 & .820 & .053 & 1.148/1.231 \\
\addlinespace[2pt]

MMLU-100
& Posthoc rationale
& 400 & .366 & .365 & .490 & .208 & .283 & 1.690/1.900 \\

MMLU-100
& Rationale first
& 400 & .430 & .425 & .608 & .300 & .308 & 1.528/1.730 \\

MMLU-100
& Rationale first + shuffled
& 400 & .424 & .435 & .528 & .258 & .270 & 1.638/2.023 \\
\bottomrule
\end{tabularx}
\end{table}

The full-manifest results support a narrow propagation claim rather than a general rationale-quality claim. On ARC-295, final correctness varies only modestly across conditions: $0.877$ for posthoc rationales, $0.880$ for rationale first, and $0.887$ for rationale first with shuffled rationales. The shuffled control also has the lowest wrong-consensus rate, $0.053$, despite breaking the semantic association between rationales and their answer sources. Its behavioral-consensus rate is lower, however, indicating that the textual control changes both convergence and correctness rather than supplying a uniformly better explanation signal.

MMLU-100 shows a stronger dependence on generation stage. Rationale first yields the highest final correctness, $0.430$, and the highest behavioral- and correct-consensus rates, $0.608$ and $0.300$. It also yields the highest wrong-consensus rate, $0.308$. Shuffled rationale first remains close in final correctness at $0.424$, while reducing behavioral consensus to $0.528$ and wrong consensus to $0.270$. Posthoc rationale produces substantially less consensus and lower final correctness.

Together, the two tasks show that generated textual context changes the strength and direction of answer propagation. Rationale-first interaction can increase convergence and correct consensus, particularly on MMLU-100, but it can also consolidate wrong answers. The strong shuffled-rationale results prevent attributing these changes solely to semantically useful explanations. We therefore interpret rationale timing and content association as propagation variables, not as a generally reliable method for improving collective accuracy. Because these rows pool routing policies and use one seed--temperature setting, they do not establish a universal ordering among rationale conditions or controllers.

\section{Closed-interface behavioral details}
\label{app:closed_behavioral_check}

Table~\ref{tab:closed_behavioral_check_app} summarizes the closed-interface checks. These results are label-observable only and are not used for the main state-space claims.

\begin{table}[h!]
\caption{Closed-interface behavioral extrapolation.}
\label{tab:closed_behavioral_check_app}
\centering
\scriptsize
\setlength{\tabcolsep}{5pt}
\begin{tabularx}{\textwidth}{p{0.22\textwidth}p{0.22\textwidth}Y}
\toprule
Model / interface & Reliable observables & Main behavioral pattern \\
\midrule
GPT-4o-mini / OpenAI API & Sampled labels, label components, label energy & Prior probe dominant label \texttt{w3} with $110/240$ samples; $H=0$ history sweep does not reach final consensus across tested temperatures; all $H=3$ and $H=10$ settings reach final behavioral consensus. \\
Gemma-3-27B / OpenRouter & Sampled labels, label components, label energy & Many settings collapse immediately to \texttt{w0:24}, including several $H=0$ settings; highest-temperature cases show residual disagreement. \\
\bottomrule
\end{tabularx}
\end{table}

\section{Baseline and history-ablation controls}
\label{app:baseline_controls}

The control suite tests whether stationary priors, generic local averaging, prompt length, or an unstructured bridge budget can reproduce the graph and memory effects. The randomized bridge-budget examples below and the matched schedule controls in Supplementary Sec.~\ref{app:matched_schedule_decomposition} serve different roles. The former match the component-bridge controller's primary-pair budget in selected settings, whereas the latter replay every threshold schedule across the full 189-trajectory mixed-population grid. Table~\ref{tab:baseline_control_summary_app} reports fixed-prior, bounded-confidence, and randomized bridge-budget controls. Fixed-prior agents sample independently from stationary distributions without prompt memory or updating. The bounded-confidence baseline uses a non-LM naming-game process with local distributional updates. Both remain fragmented in the tested configuration.

The randomized bridge-budget control matches the standard component-bridge controller's number of primary pairs but selects shuffled cross-component edges before well-mixed fallback. It can sometimes reach behavioral and score-state connectedness when social evidence is retained, showing that structured bridge routing is not uniquely capable of repair. These controls are representative rather than a matched multi-seed controller grid, so they establish possible alternative mechanisms without supplying a comparative stability estimate.

\begin{table}[h]
\caption{Baseline controls. $C_{\mathrm{label}}/C_{\mathrm{state}}$ gives final label/state component counts when state components are defined; ``Full frac.'' is the fraction of rounds with full behavioral consensus.}
\label{tab:baseline_control_summary_app}
\centering
\scriptsize
\setlength{\tabcolsep}{4pt}
\renewcommand{\arraystretch}{1.12}
\begin{tabularx}{\textwidth}{@{}p{0.23\textwidth}p{0.18\textwidth}Xcc@{}}
\toprule
Control & Setting & Final labels & $C_{\mathrm{label}}/C_{\mathrm{state}}$ & Full frac. \\
\midrule
Fixed-prior agents
& $N=10,M=10$
& \texttt{w0:1, w2:2, w3:2, w5:1, w6:1, w7:2, w9:1}
& 7/--
& 0.000 \\
Bounded-confidence naming game
& $N=10,M=10$
& \texttt{w0:1, w2:2, w3:2, w5:1, w6:1, w7:2, w9:1}
& 7/--
& 0.000 \\
Random bridge-budget
& $H=0,\tau=0.1$
& \texttt{cal:3, fel:2, pel:3, tal:2}
& 4/8
& 0.000 \\
Random bridge-budget
& $H=3,\tau=0.1$
& \texttt{vel:9, sol:1}
& 2/2
& 0.490 \\
Random bridge-budget
& $H=3,\tau=0.3$
& \texttt{vel:10}
& 1/1
& 0.710 \\
Random bridge-budget
& $H=10,\tau=0.7$
& \texttt{sol:10}
& 1/1
& 0.805 \\
\bottomrule
\end{tabularx}
\end{table}

Table~\ref{tab:history_ablation_controls_app} summarizes history controls at $H=3,\tau=0.3$. The partner-label-hidden condition preserves the number of history records, the agent's own labels, the match/mismatch outcome, and all reward text; it replaces only each partner label with the fixed placeholder \texttt{[hidden]}. The resulting prompt is not exactly length matched: at $H=3$ it is six tokens longer under every tokenizer used in the mixed population, because \texttt{[hidden]} occupies more tokens than a tokenizer-safe nonce label. Its loss of coordination therefore cannot be attributed to receiving a shorter prompt.

Shuffling records within an agent or across agents preserves token length exactly. Global history has the same length after the history window is full, but can contain more records during the initial rounds. These controls show that explicit partner-label identity and the organization of retained social evidence affect coordination; they do not rely on a shorter-prompt comparison.

Table~\ref{tab:baseline_control_summary_app} rules out two simple explanations for the main results. Fixed-prior sampling remains fragmented, so the repeated-game consensus results are not explained by stationary label priors alone. The bounded-confidence naming-game baseline also remains fragmented in the tested configuration, so generic local distributional averaging does not reproduce the observed repair behavior. The randomized bridge-budget baseline gives a more nuanced control: with $H=0$, targeting the standard matcher's per-round bridge-pair count is still insufficient, while with retained history it can sometimes reach behavioral consensus. These representative repaired cases show that randomized cross-basin exposure can be sufficient in some retained-memory settings. The larger matched controller grids nevertheless show that repair reliability, terminal persistence, and state-space cleanliness vary with the bridge objective. Thus, the supported claim is not that structured routing is uniquely capable of repair, but that runtime graph design affects how reliably and cleanly repair occurs.

\begin{table}[p]
\caption{History-ablation controls at $H=3,\tau=0.3$. ``Stable'' indicates full behavioral consensus throughout the final 20 rounds, and ``Full frac.'' is the fraction of rounds with full behavioral consensus.}
\label{tab:history_ablation_controls_app}
\centering
\tiny
\setlength{\tabcolsep}{4pt}
\renewcommand{\arraystretch}{1.12}
\begin{tabularx}{\linewidth}{@{}p{0.15\linewidth}p{0.14\linewidth}p{0.18\linewidth}c c c X@{}}
\toprule
Setting & History ablation & Final labels & $C_{\mathrm{label}}/C_{\mathrm{state}}$ & Stable & Full frac. & Takeaway \\
\midrule
Qwen2.5-7B well-mixed & Global history & \texttt{zel:10} & 1/1 & Yes & 0.995 & Shared convention evidence is sufficient in the homogeneous case. \\
Qwen2.5-7B well-mixed & Shuffled within agent & \texttt{zel:10} & 1/1 & Yes & 0.950 & Exact temporal order is not essential. \\
Qwen2.5-7B well-mixed & Shuffled across agents & \texttt{zel:10} & 1/1 & No & 0.550 & Consensus can occur, but stability weakens. \\
Qwen2.5-7B well-mixed & Partner label hidden & \texttt{kov:4, zel:6} & 2/2 & No & 0.000 & Memory without partner feedback does not repair fragmentation. \\
Mixed well-mixed & Global history & \texttt{sol:10} & 1/2 & No & 0.905 & Behavioral consensus can emerge with residual state structure. \\
Mixed well-mixed & Shuffled within agent & \texttt{sol:10} & 1/2 & No & 0.835 & Local order is less important than retained convention evidence. \\
Mixed well-mixed & Shuffled across agents & \texttt{vel:10} & 1/2 & No & 0.405 & Cross-agent history can coordinate labels but weakens stability. \\
Mixed well-mixed & Partner label hidden & \texttt{fel:4, mur:1, sol:3, vel:2} & 4/4 & No & 0.000 & Hiding explicit partner-label identity eliminates repair in this control. \\
Mixed component bridge & Global history & \texttt{sol:10} & 1/2 & Yes & 0.940 & Bridge routing plus global convention evidence repairs behaviorally. \\
Mixed component bridge & Shuffled within agent & \texttt{sol:10} & 1/2 & Yes & 0.845 & Repair is robust to within-agent order shuffling. \\
Mixed component bridge & Shuffled across agents & \texttt{sol:10} & 1/2 & No & 0.830 & Behavioral consensus persists, but terminal stability weakens. \\
Mixed component bridge & Partner label hidden & \texttt{cal:1, gar:2, mur:2, sol:2, tal:1, vel:2} & 6/6 & No & 0.000 & Bridge exposure cannot help if prompts omit rendered partner-label evidence. \\
Mixed label disagreement & Global history & \texttt{sol:10} & 1/1 & Yes & 0.955 & Surface disagreement routing plus global history gives clean repair here. \\
Mixed label disagreement & Shuffled within agent & \texttt{sol:10} & 1/2 & Yes & 0.945 & Behavioral repair survives order shuffling but leaves state residue. \\
Mixed label disagreement & Shuffled across agents & \texttt{sol:10} & 1/2 & No & 0.780 & Behavioral repair persists but is less stable. \\
Mixed label disagreement & Partner label hidden & \texttt{cal:1, gar:2, mur:2, sol:2, tal:1, vel:2} & 6/6 & No & 0.000 & Removing partner-label evidence eliminates repair in this control. \\
\bottomrule
\end{tabularx}
\end{table}

\begin{figure}[h!]
    \centering
    \includegraphics[width=0.82\textwidth]{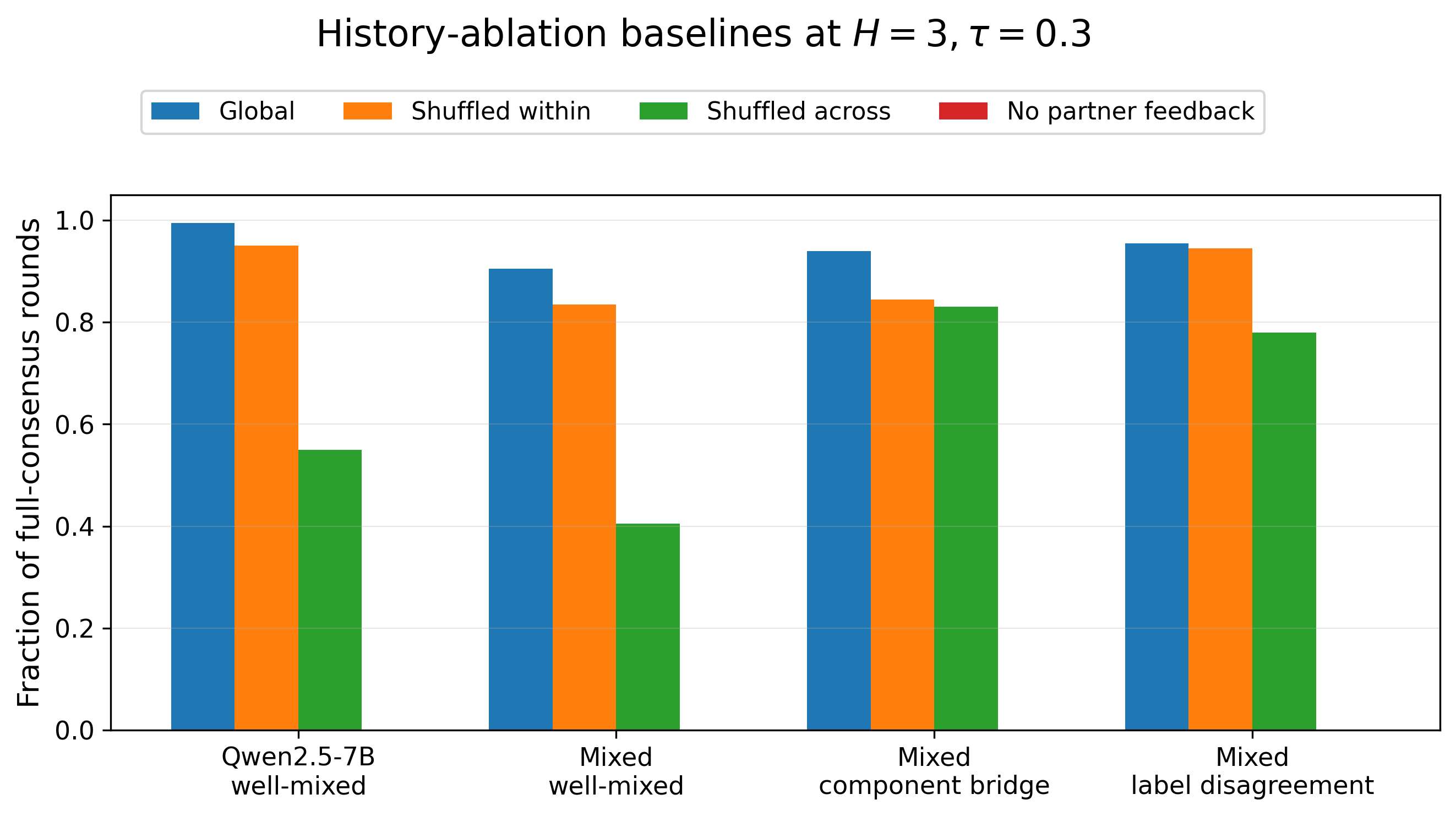}
    \caption{Fraction of full-consensus rounds for history-ablation controls at $H=3,\tau=0.3$. Shuffled and global histories often preserve behavioral consensus, while removing rendered partner-label evidence eliminates full-consensus rounds across all tested settings.}
    \label{fig:history_ablation_full_consensus_fraction_app}
\end{figure}

\FloatBarrier

\section{Trajectory and prompt diagnostics}
\label{app:additional_diagnostics}

\subsection{Counterfactual prompt-sensitivity probes}
\label{app:counterfactual_influence}

When exact restricted score-state distributions are available, we can estimate exploratory counterfactual prompt sensitivity by ablating one agent's rendered contribution from another agent's prompt. If $C_i^t$ is agent $i$'s prompt context and $m_j^t$ is the contribution from agent $j$, define
\begin{equation}
    \hat a^{\mathrm{cf}}_{ij}(t)=
    \JS\!\left(
    \hat s_i(\cdot\mid C_i^t),
    \hat s_i(\cdot\mid C_i^t\setminus m_j^t)
    \right).
    \label{eq:cf_influence}
\end{equation}
This quantity is directed and measures prompt sensitivity, not adoption. It is not used as main evidence for the bridge-repair claims. Table~\ref{tab:cf_influence_condensed_app} reports representative probes. Within/cross averages are computed using final label components; when the final label graph has a single component, the cross-component average is undefined and reported as ``--''.

\begin{table}[h]
\caption{Representative counterfactual prompt-sensitivity probes. The weights are directed JS-divergence sensitivities between restricted score-state distributions under the full prompt and under a prompt with one rendered source contribution removed. Within/cross averages use final label components; cross averages are undefined when the final label graph has one component.}
\label{tab:cf_influence_condensed_app}
\centering
\scriptsize
\setlength{\tabcolsep}{5pt}
\renewcommand{\arraystretch}{1.12}
\begin{tabularx}{\textwidth}{@{}llXccc@{}}
\toprule
Population & Setting & Final labels & Label/State comps. & Mean CF & Within/Cross CF \\
\midrule
Qwen2.5-7B & $H=0,\tau=0.1$ & \texttt{dek:6, kov:4} & 2/2 & 0.111 & 0.238/0.000 \\
Qwen2.5-7B & $H=3,\tau=0.3$ & \texttt{zel:10} & 1/1 & 0.111 & 0.111/-- \\
Mixed four-model & $H=3,\tau=0.1$ & \texttt{vel:5, sol:5} & 2/2 & 0.022 & 0.014/0.029 \\
Mixed four-model & $H=3,\tau=0.3$ & \texttt{vel:10} & 1/2 & 0.0105 & 0.0105/-- \\
\bottomrule
\end{tabularx}
\end{table}

\FloatBarrier

\subsection{Residual score structure and perturbation stability}
\label{app:latent_residue}

We branch from behavioral-consensus states and classify the pre-branch score graph as clean when it has one state component and residual when sampled consensus coexists with multiple state components. Each qualifying state produces an unmodified continuation and a matched continuation in which one selected agent is forced once to emit a challenger label. This campaign therefore supports residual score structure as a warning signal, not a completed causal claim that such structure amplifies perturbations.

\begin{table}[h]
\caption{Exploratory latent-residue perturbation summary. ``Control loss'' reports whether the unmodified continuation ever leaves full behavioral consensus. ``Added disagreement'' is the mean number of additional non-consensus rounds induced by the matched one-agent intervention.}
\label{tab:latent_residue_app}
\centering
\small
\setlength{\tabcolsep}{6pt}
\begin{tabular}{lcccc}
\toprule
Pre-state & $n$ & Control loss & Control final cons. & Added disagreement \\
\midrule
Clean & 73 & 7/73 & 72/73 & 1.08 \\
Residual & 8 & 5/8 & 6/8 & 1.25 \\
\bottomrule
\end{tabular}
\end{table}

Residual controls leave consensus more frequently, suggesting that latent residue is associated with intrinsically fragile regimes. The matched intervention, however, adds only a small amount of disagreement, recovery is typically immediate, and no convention switch occurs. This campaign therefore supports residue as a warning signal, not a completed causal claim that residue amplifies perturbations.

\FloatBarrier

\subsection{Audited early-window predictive diagnostics}
\label{app:predictive_diagnostics}

The audited predictive dataset contains one $K=10$ early-window summary row per saved run, using rounds 0--10 inclusive, for 99 unique runs. Targets are final behavioral consensus, final score-state connectedness, and stable final-20 behavioral consensus. Positive/negative class counts are 20/79, 17/82, and 18/81, respectively.

Feature groups are metadata only, graph only, and energy plus graph. The metadata baseline includes graph mode, history horizon, temperature, similarity threshold, and inferred population type. Numeric variables are median-imputed and standardized; categorical variables are most-frequent-imputed and one-hot encoded. All preprocessing is fitted inside each training fold. Classifiers are balanced logistic regressions with no hyperparameter search. Evaluation uses five-fold \texttt{GroupKFold} grouped by run identifier, and each reported AUROC is computed from the pooled out-of-fold positive-class probabilities rather than averaged across folds.

\begin{table}[h]
\caption{Audited pooled out-of-fold AUROC using rounds 0--10. Values measure within-grid discriminability under the sampled controller, model, history, temperature, and population regimes; they do not test extrapolation to unseen regimes.}
\label{tab:prediction_audited_app}
\centering
\small
\setlength{\tabcolsep}{6pt}
\begin{tabular}{lccc}
\toprule
Target & Metadata & Graph & Energy+graph \\
\midrule
Final behavioral consensus & 0.953 & 0.983 & 0.997 \\
Final score-state connectedness & 0.961 & 0.993 & 0.982 \\
Stable full consensus & 0.966 & 0.958 & 0.991 \\
\bottomrule
\end{tabular}
\end{table}

An explicit audit reproduces all nine values to three decimals and finds no target column, final-summary field, file path, or run identifier in the reported feature matrices. The high metadata AUROC is expected because the sampled policies, histories, and populations have sharply different outcome rates. Grouping by run prevents literal duplication, but related sweep configurations can still appear in different folds; this is therefore a within-grid warning diagnostic rather than held-out-controller, held-out-family, or external validation.

Two archived code paths generate the stable-consensus target: saved summaries use the final-20 behavioral definition, while a builder fallback uses label-plus-state persistence from first full consensus. Recomputing a uniform final-20 behavioral target changes 0 of the 99 labels, so the reported AUROCs do not change. No uncertainty interval was computed for these values.

\section{Reproducibility details}
\label{app:reproducibility_details}

The public implementation and paper-facing reproduction scripts are available at
\url{https://github.com/cedar-lau/llm-graph-control}.

\paragraph{Hardware and fixed population ordering.}
Experiments whose largest checkpoint had at most 7B parameters were executed on two NVIDIA RTX 3090 GPUs. The large-model and large heterogeneous-population experiments were executed on two NVIDIA A100 GPUs.

The central mixed population has $N=10$ and uses the fixed ordered checkpoint sequence
\[
\begin{split}
(&\text{Qwen2.5-7B},\text{Qwen2-7B},\text{Yi-6B},\text{TinyLlama},\\
 &\text{Qwen2.5-7B},\text{Qwen2-7B},\text{Yi-6B},\text{TinyLlama},\\
 &\text{Qwen2.5-7B},\text{Qwen2-7B}).
\end{split}
\]
The resulting multiplicities are three Qwen2.5-7B agents, three Qwen2-7B agents, two Yi-1.5-6B agents, and two TinyLlama-1.1B agents. The same sequence is recorded across the raw threshold grid, the active-set, pair-count, and full-coverage controls, prior-normalized routing, and the mixed-population history controls.

The principal large heterogeneous cohort uses the ordered sequence
\[
(\text{32B},\text{14B},\text{Yi},\text{TinyLlama},
 \text{32B},\text{14B},\text{Yi},\text{TinyLlama},
 \text{32B},\text{14B}),
\]
giving multiplicities $3,3,2,2$.

\paragraph{Loading, scoring, and sampling.}
Open-weight models were loaded in bfloat16 with \texttt{device\_map="auto"} and no quantization configuration. Tokenizer identifiers equal the corresponding model identifiers. Each label decision uses a single forward pass rather than free-form generation. The model's first-token logits are restricted to the tokenizer-safe allowed labels. A temperature-one softmax over these logits defines the reported score state, while the run temperature $\tau$ defines the categorical distribution from which the emitted label is sampled. Consequently, \texttt{top\_p}, \texttt{top\_k}, and repetition penalties do not enter the naming-game or label-only task-choice decisions.

The resolved configurations record \texttt{max\_new\_tokens=4}, but this field is not used on the restricted-label path. Each scoring forward pass has batch dimension one. The rationale-sharing campaign uses a separate free-form generation path with \texttt{max\_new\_tokens=64}, temperature $0.2$, and sampling enabled. No \texttt{top\_p} or \texttt{top\_k} override is supplied on that path.

\paragraph{Exact naming-game prompt and reward.}
The naming-game experiments use the following system and user prompt templates. The system message is:

\begin{quote}
\texttt{Answer with exactly one token from the allowed list. No extra words.}
\end{quote}

The following is a representative scored user message with three retained records; the allowed-label order is query specific.

{\small
\begin{verbatim}
Goal: coordinate by replying with the SAME name as your partner.
Your objective is to maximize coordination with your current partner.
Your recent interactions (most recent first):
Partner: sol | You: vel | Outcome: mismatch | Reward: 0
Partner: fel | You: mur | Outcome: mismatch | Reward: 0
Partner: tal | You: cal | Outcome: mismatch | Reward: 0
Your last reward: 0
Your cumulative reward: 0
Allowed names: cal, pel, lod, mur, lim, gar, fel, vel, tal, sol
Reply with exactly one token from the allowed list to maximize coordination.
\end{verbatim}
}

The model-specific tokenizer chat template wraps this user message and the system message with \texttt{add\_generation\_prompt=true}. History records are shown most recent first. At $H=0$, the history block is the literal text \texttt{(no recent interactions)}. The executed prompt contains no round-index line.

For the reported matching objective, paired agents $i$ and $j$ receive the symmetric round reward
\[
r_i(t)=r_j(t)=
\mathbf{1}\{y_i(t)=y_j(t)\}.
\]
Each cumulative reward is initialized to zero and updated after the round. Both the most recent reward and cumulative reward are included in subsequent prompts. Rewards affect prompt text only; routing does not read them. The reported experiments assign rewards of $1$ for a match and $0$ for
a mismatch.

The allowed-label order is independently reshuffled for each agent and round using a deterministic generator seeded by
\[
(\mathrm{seed}+1)\,1000003 + 10007\,t+i,
\]
where $t$ is the round and $i$ is the agent index. The central mixed-population label set is
\[
\{\texttt{cal},\texttt{pel},\texttt{lod},\texttt{mur},
\texttt{lim},\texttt{gar},\texttt{fel},\texttt{vel},
\texttt{tal},\texttt{sol}\}.
\]

\paragraph{History-control implementation.}
The partner-label-hidden condition replaces every rendered partner label with \texttt{[hidden]} while preserving the agent's own label, the number of records, the match/mismatch field, and the reward. Thus, on matched records the partner label remains inferable from the outcome; this condition removes explicit partner-label identity rather than all partner feedback. Shuffled-within-agent uses the same local records in a different order. Shuffled-across-agents samples records from other agents' histories. Global history takes the most recent records from the pooled population history. These transformations affect only a cloned prompt history and do not modify the stored trajectory state.

At $H=3$, replacing the three one-token partner labels with \texttt{[hidden]} adds exactly six tokens under each of the four tokenizers in the mixed population. The partner-label-hidden prompt is therefore slightly longer, not shorter, than the corresponding local history prompt. The within-agent and across-agent order controls are exactly length matched. Global history becomes length matched once the history window is full.

\paragraph{Task manifests and item selection.}
The label-only ARC-Challenge and MMLU experiments use homogeneous populations of six Qwen2.5-7B-Instruct agents for 20 rounds, with $H=3$, $\Delta=0.25$, temperatures $\{0.3,0.7\}$, seeds $\{11,12,13\}$, and answer labels $\{\texttt{A},\texttt{B},\texttt{C},\texttt{D}\}$. No rationale is generated in either label-only campaign. Every run within a task uses the same saved manifest.

MMLU subjects are pooled rather than sampled with per-subject quotas; 36 subjects appear in the final manifest. Qwen2.5-7B-Instruct scores the candidate questions at $H=0$, and items are ordered by decreasing entropy of the four-choice score distribution. Ties are resolved by increasing top-two margin and then increasing top-one probability. An ambiguity pre-filter admitted only 51 candidates, fewer than the requested 100, so the implementation fell back to the complete 1000-item pool; the filter thresholds therefore do not affect the final manifest. The scoring model's top-one accuracy is $0.690$ on the candidate pool and $0.380$ on the selected set. ``Hard'' consequently means high predictive uncertainty for this scoring model, not a human difficulty annotation.

The matched schedule suite contains 378 control trajectories: 189 active-set-matched random trajectories and 189 pair-count-matched random trajectories. Each control is linked to one of the 189 threshold references through a saved reference identifier and resolved configuration hash. The active-set policy preserves the complete per-round participating-agent set, while the pair-count policy preserves the per-round number of pairs. Round-level schedule audits verify the required invariants for all 200 rounds of every trajectory. Endpoint and participation summaries are read from the saved control artifacts. Paired uncertainty intervals use configuration-cluster bootstrap resampling over the 42 retained-memory $(H,\tau,\Delta)$ configurations, with all three seeds retained together inside each resampled configuration.

The prior-normalized routing suite contains 360 stationary-prior probes and 243 population trajectories: 189 threshold-similarity trajectories, 27 state-component trajectories, and 27 state-distance trajectories. Each of the four model families contributes 90 prior probes, consisting of 30 randomized allowed-label orders under each of seeds 11--13. Probe scores are remapped to the canonical label order before the model-specific geometric-mean prior is constructed. Every population trajectory records the consumed prior-bundle hash, normalization formula, routing-state source, sampling-state source, and no-lookahead flag. Raw restricted logits remain the source of sampled outputs, while prior-normalized states affect routing only. All 243 trajectories contain 200 rounds and pass the suite-level configuration, completion, and score-path validations. Both raw-state and normalized-state endpoint diagnostics are retained.

The population-composition study contains 150 runs across five cohorts, five routing policies, two history conditions, and three seeds. Cohort outcomes are read from per-run summaries, while transient and stable consensus are recomputed from round traces. The Qwen2.5-14B/32B attainment-and-persistence comparison uses all 594 homogeneous trajectories, constructs 297 exact setting pairs, and reports the 72 retained-memory non-threshold settings per model as its principal subset. The homogeneous Yi/TinyLlama suite contains 594 trajectories, 297 per model: 27 well-mixed history settings, 189 threshold-similarity settings, and 27 settings under each of the state-component, label-disagreement, and state-distance bridge policies. All 594 expected setting trajectories are present. The susceptibility study contains 450 stationary-prior probes, 900 non-empty synthetic-history probes, and 12 seeded-population runs. The predictive dataset contains 99 unique runs and one reported $K=10$ row per run.

The canonical manuscript-reproduction path consists of the paper-facing entrypoints in \path{scripts/open/}, the installable experiment package in \path{switched_llm/experiments/open/}, and the top-level \path{scripts/run_open_*.sh} orchestration wrappers. The \path{switched_llm/} package also contains the model backends, graph utilities, measurements, prompt renderers, task loaders, and reusable simulation logic. Baseline and history-ablation implementations are located in \path{scripts/baselines/}, with paper-scale wrappers in \path{scripts/run_baselines_*.sh}. The generic YAML runner \path{scripts/run_experiment.py} remains available for configuration-driven experiments, but manuscript reproduction uses the paper-facing open-weight entrypoints and wrappers.

The task-grounded suite is implemented by \path{scripts/open/task_choice_coordination.py}, with item selection in \path{scripts/open/select_task_items.py}. Paper-facing task tables and answer-propagation analyses are produced by \path{scripts/open/make_task_suite_tables.py}, \path{scripts/open/analyze_task_answer_shifts.py}, \path{scripts/open/analyze_task_directional_transitions.py}, and \path{scripts/open/analyze_task_rationale_effects.py}. Packaged export helpers under \path{scripts/open/} produce the manuscript-facing archives and tabular summaries.

Fresh experiments write under \path{outputs/}, normally through an explicit \texttt{--output-dir}. This directory is excluded from version control because paper-scale state and round traces can be large. Full experiment artifacts are distributed separately as supplementary archives or release assets, while \path{sample_outputs/} contains only small schema examples. A completed run commonly contains \path{summary.json}, \path{round_logs.jsonl}, \path{state_vectors.jsonl}, and, when enabled, a \path{graph_artifacts/} directory containing graph matrices, spectra, and derived summaries. Sweep-level aggregation produces \path{aggregate_summary.csv} and \path{aggregate_summary.json}.

All paper-facing tables and figures are generated from saved artifacts rather than manual transcription. Final metrics are read from per-run summaries, while stable consensus, transient consensus, full-consensus fractions, and other temporal quantities are recomputed from round-level traces. State-energy checks are recomputed from the restricted score-state vectors in \path{state_vectors.jsonl} and agree with the corresponding saved round-level measurements up to numerical precision.

Coverage statistics are computed from the realized per-round matchings, while score-state concentration statistics are computed from all pairwise Jensen--Shannon distances between the restricted score distributions. Configuration-level summaries group runs by population, controller, history, temperature, and threshold, with seed treated as the replication axis.

Table~\ref{tab:homogeneous_history_all_models_app} uses the well-mixed TinyLlama, Yi, and Qwen2.5 history aggregates, pooling seeds $11$--$13$ and temperatures
$\tau\in\{0.1,0.3,0.7\}$ within each $H$. Tables~\ref{tab:model_scale_history_app} and \ref{tab:model_scale_controller_app} retain the detailed Qwen2.5-14B/32B stability and controller statistics, while Table~\ref{tab:yi_tiny_controller_app} reports the complete TinyLlama/Yi controller grids. Retained controller rows pool $H\in\{3,10\}$ at $\Delta=0.25$, and threshold-similarity rows retain the full seven-threshold grid. Tables~\ref{tab:population_size_stress_app} and \ref{tab:vocab_robustness_app} are likewise generated from per-setting summaries and round-level recomputation where required. 

Directional cross-family adoption statistics use the no-14B cohort at $H=3$, $\tau=0.3$, and $\Delta=0.25$ under state-component, label-disagreement, and well-mixed routing, pooled over seeds 11--13. A counted directed event requires that target agent $i$ observed source agent $j$ at round $t$, that $y_i(t)\neq y_j(t)$, and that adoption is evaluated by whether
\[
    y_i(t+1)=y_j(t).
\]
Each unordered interaction can therefore contribute two directed events. Table~\ref{tab:family_adoption_app} reports exact pooled qualifying-event counts separately for Yi agents adopting a non-Yi source label and non-Yi agents adopting a Yi source label under each policy. 

Every stochastic run records its random seed and resolved configuration in its saved metadata. Representative trajectory examples use seed 11 unless otherwise stated, while the multi-seed controller, scale, robustness, and population-composition contrasts use seeds 11--13. Saved summaries and configuration metadata identify the model, population composition, graph mode, history horizon, temperature, state threshold, population size, vocabulary, and horizon length.

Python, NumPy, and Torch CPU and CUDA random generators are initialized from the saved run seed. Deterministic-algorithm flags were not enabled, so the configurations support statistical reproduction but do not guarantee bitwise identity across different GPU, driver, or library versions. Python, NumPy, and Torch random generators were initialized from each run's saved seed. Deterministic-algorithm flags were not enabled, so the reported configurations support statistical reproduction but do not guarantee bitwise-identical trajectories across hardware and software environments. The code release specifies the tested software dependencies and execution requirements.

\paragraph{Runtime bridge matching and well-mixed fallback.}
For bridge-policy runs, the controller first constructs the primary bridge candidate graph from previous-round observations only. It then selects primary pairs by applying a greedy high-weight maximal matching to the positive-weight candidate edges: candidate edges are sorted by weight with a small random tie-breaker, and an edge is accepted only if neither endpoint has already been matched. The reported bridge-controller runs use \texttt{bridge\_fallback=well\_mixed}. Under this fallback, agents not selected by the primary bridge matching are shuffled uniformly at random and paired consecutively; if the unmatched set has odd size, one agent can remain unpaired in that round. The fallback is applied only to unmatched agents and does not replace or rewire endpoints already selected by the primary bridge graph.

In the reported state-component bridge runs, the primary bridge graph is built from previous state-similarity components: candidate edges connect agents in different previous state components, with the component threshold given by the run's \(\Delta\) or by the configured \texttt{component\_threshold}. When \texttt{bridge\_weight=distance}, these primary candidate edges are weighted by previous-round Jensen--Shannon distance. Label-disagreement routing instead builds primary candidate edges between agents with different previous sampled labels, and state-distance bridge routing builds primary candidate edges whose previous-round state distance lies in the configured bridge band. Round-level metadata records \texttt{num\_pairs\_from\_bridge\_graph}, \texttt{num\_pairs\_from\_fallback}, \texttt{fraction\_pairs\_from\_bridge\_graph}, and the corresponding cross-label and cross-state-component pair fractions. Accordingly, the bridge-policy rows represent bridge-priority controllers with well-mixed completion rather than pure bridge-only matchings in every round.

\FloatBarrier

\section{Proof sketches and technical details}
\label{app:proofs}

Throughout this section, the interpretive consensus reference dynamics are
\begin{equation}
    \dot{x}(t)
    =
    -\bigl(L(t)\otimes I_M\bigr)x(t)+r(t),
    \label{eq:reference_dynamics_app}
\end{equation}
where $L(t)$ is the Laplacian of an undirected reference graph. These dynamics are used to motivate diagnostics and are not assumed to be the update rule implemented by the language models.

\subsection{Centroid and disagreement decomposition}

Let $\bar{x}(t)=N^{-1}\sum_i x_i(t)$ and $\delta(t)=(\Pi_N\otimes I_M)x(t)$. For the undirected Laplacians considered here, $L(t)\one=0$ and $\one^\top L(t)=0$,
\begin{equation}
    \dot{\bar{x}}(t)=\frac{1}{N}\sum_{i=1}^N r_i(t).
\end{equation}
Moreover, $\Pi_NL(t)=L(t)\Pi_N=L(t)$ for undirected Laplacians, so
\begin{equation}
    \dot{\delta}(t)=-(L(t)\otimes I_M)\delta(t)+(\Pi_N\otimes I_M)r(t).
\end{equation}
If all agents share the same drift $r_i(t)=r_c(t)$, then $r(t)=\one\otimes r_c(t)$ and $(\Pi_N\otimes I_M)r(t)=0$.

\subsection{Common disagreement Lyapunov function}

For $V(\delta)=\frac12\|\delta\|_2^2$ and $r\equiv0$,
\begin{equation}
    \dot V(\delta)=-\delta^\top(L(t)\otimes I_M)\delta.
\end{equation}
Because $\delta$ lies in the orthogonal complement of the consensus manifold, the zero eigenvalue of $L(t)$ is excluded. If $\lambda_2(L(t))\geq\lambda>0$, then
\begin{equation}
    \delta^\top(L(t)\otimes I_M)\delta\geq \lambda\|\delta\|_2^2,
\end{equation}
which yields $\dot V\leq -2\lambda V$ and $\|\delta(t)\|_2\leq e^{-\lambda(t-t_0)}\|\delta(t_0)\|_2$.

\subsection{Eventual partition and conditional cluster consensus}

If after $T_\star$ the graph has no edges between clusters $C_1,\ldots,C_K$, then after reordering agents,
\begin{equation}
    L(t)
    =
    \blkdiag\!\left(
        L_{C_1}(t),\ldots,L_{C_K}(t)
    \right),
    \qquad t\geq T_\star .
\end{equation}
The reference dynamics therefore split into independent cluster subsystems. Block diagonality alone establishes decoupling but does not guarantee convergence within each cluster. If, in addition, each cluster satisfies a uniform connectivity condition
\begin{equation}
    \lambda_2\!\left(L_{C_k}(t)\right)
    \geq
    \underline{\lambda}_k>0,
    \qquad
    t\geq T_\star ,
\end{equation}
or an appropriate repeated joint-connectivity condition, then the Lyapunov argument of the preceding subsection gives exponential decay of within-cluster disagreement. Since no term couples the cluster centroids, global consensus is not implied unless the limiting cluster centroids happen to coincide.

\subsection{Type-gap edge exclusion}

The empirical state-similarity graph is defined by thresholding Jensen--Shannon distance,
\[
    a^{\mathrm{state}}_{ij}(t)
    =
    \mathbf{1}\{\DJS(s_i(t),s_j(t))\leq\Delta\}.
\]
The type-gap condition should therefore be stated in the same metric. Let $d(p,q)=\DJS(p,q)$. Since $\DJS$ is the square root of the Jensen--Shannon divergence, $d$ is a metric on probability distributions. Suppose that every agent of type $m$ remains within a $d$-ball of radius $B_m$ around a type prototype $\mu_m$:
\[
    d(s_i(t),\mu_m)\leq B_m
    \qquad
    \text{for all agents $i$ of type $m$.}
\]
For an agent $i$ of type $m$ and an agent $j$ of type $n$, the triangle inequality gives
\begin{align}
    d(s_i(t),s_j(t))
    &\geq
    d(\mu_m,\mu_n)
    -
    d(s_i(t),\mu_m)
    -
    d(s_j(t),\mu_n) \notag \\
    &\geq
    d(\mu_m,\mu_n)-B_m-B_n .
\end{align}
Thus, if
\[
    d(\mu_m,\mu_n)>\Delta+B_m+B_n,
\]
then
\[
    \DJS(s_i(t),s_j(t))>\Delta
\]
for all such cross-type pairs and all times $t$. Therefore the state-similarity threshold rule cannot activate cross-type edges between types $m$ and $n$. In the common-radius case $B_m=B_n=B$, the sufficient condition reduces to
\[
    \DJS(\mu_m,\mu_n)>\Delta+2B .
\]

\FloatBarrier

\bibliographystyle{plainnat}